\documentclass[journal]{IEEEtran}
\usepackage{amsmath,amsfonts}
\usepackage{algorithmic}
\usepackage{algorithm}
\usepackage{array}
\usepackage[caption=false,font=normalsize,labelfont=sf,textfont=sf]{subfig}
\usepackage{textcomp}
\usepackage{stfloats}
\usepackage{url}
\usepackage{verbatim}
\usepackage{graphicx}
\usepackage{cite}
\usepackage{booktabs}
\usepackage{multirow}
\usepackage{color}
\begin{document}

\title{DSDFormer: An Innovative Transformer-Mamba Framework for Robust High-Precision Driver Distraction Identification}

\author{Junzhou Chen, Zirui Zhang, Jing Yu, Heqiang Huang, Ronghui Zhang, Xuemiao Xu,\\ Bin Sheng, Hong Yan, \textit{Fellow, IEEE}

\thanks{This work has been submitted to the IEEE for possible publication. Copyright may be transferred without notice, after which this version may no longer be accessible.}
\thanks{Junzhou Chen, Zirui Zhang, Jing Yu, Heqiang Huang, Ronghui Zhang are with Guangdong Provincial Key Laboratory of Intelligent Transportation System, School of Intelligent Systems Engineering, Sun Yat-Sen University, Guangzhou 510275, China (e-mail: chenjunzhou@mail.sysu.edu.cn; zhangzr23@mail2.sysu.edu.cn; yujing68@mail2.sysu.edu.cn;\\ huanghq77@mail2.sysu.edu.cn; zhangrh25@mail.sysu.edu.cn). \\ \textit{(Corresponding author: Ronghui Zhang.)}}
\thanks{Xuemiao Xu is with the School of Computer Science and Engineering at South China University of Technology (e-mail: xuemx@scut.edu.cn).}
\thanks{Bin Sheng is with the Department of Computer Science and Engineering, Shanghai Jiao Tong University (e-mail: shengbin@sjtu.edu.cn).}
\thanks{Hong Yan is with the Department of Electrical Engineering, City University of Hong Kong, Kowloon, Hong Kong (e-mail: h.yan@cityu.edu.hk).}}



\maketitle

\begin{abstract}

Driver distraction remains a leading cause of traffic accidents, posing a critical threat to road safety globally. As intelligent transportation systems evolve, accurate and real-time identification of driver distraction has become essential. However, existing methods struggle to capture both global contextual and fine-grained local features while contending with noisy labels in training datasets. To address these challenges, we propose DSDFormer, a novel framework that integrates the strengths of Transformer and Mamba architectures through a Dual State Domain Attention (DSDA) mechanism, enabling a balance between long-range dependencies and detailed feature extraction for robust driver behavior recognition. Additionally, we introduce Temporal Reasoning Confident Learning (TRCL), an unsupervised approach that refines noisy labels by leveraging spatiotemporal correlations in video sequences. Our model achieves state-of-the-art performance on the AUC-V1, AUC-V2, and 100-Driver datasets and demonstrates real-time processing efficiency on the NVIDIA Jetson AGX Orin platform. Extensive experimental results confirm that DSDFormer and TRCL significantly improve both the accuracy and robustness of driver distraction detection, offering a scalable solution to enhance road safety.

\end{abstract}

\begin{IEEEkeywords}
driver distraction identification, Mamba, transformers, confident learning, traffic accidents.
\end{IEEEkeywords}

\section{Introduction}
\IEEEPARstart{R}{oad} traffic accidents have increased in frequency, leading to severe injuries and significant property losses. In 2020, road traffic accidents in the United States resulted in 38,824 fatalities, 2.28 million injuries, and direct economic losses of 340 billion US dollars \cite{data1}. Driver distraction is a major factor in these accidents. In 2019, distracted driving caused 10,546 deaths, 1.3 million injuries, and property damage totaling 98.2 billion US dollars \cite{data2}. Reducing driver distractions to improve road safety is crucial, especially with the growth of smart cities and intelligent transportation systems (ITS).

Smart cities and ITS, as future pathways of urban development, aim to leverage artificial intelligence (AI), the Internet of Things (IoT), and big data analytics to enhance transportation safety, efficiency, pollution control, and other municipal services. However, the proliferation of intelligent devices in vehicles has increased driving tasks and mental workload for drivers. Thus, there is an urgent need to detect driver distraction behaviors and provide proactive warnings to enhance road safety. Developing an efficient and accurate algorithm for driver distraction detection is a significant challenge. Contact-based methods, which monitor vital signs like blood pressure, pulse, and respiration, could disrupt the driver’s normal performance. In contrast, vision-based driver distraction detection offers a promising non-contact solution with a single in-vehicle camera. This approach holds great potential for implementation in the rapidly evolving landscape of smart cities and ITS.

\begin{figure*}[!t]
\centering
\includegraphics[width=6.5in]{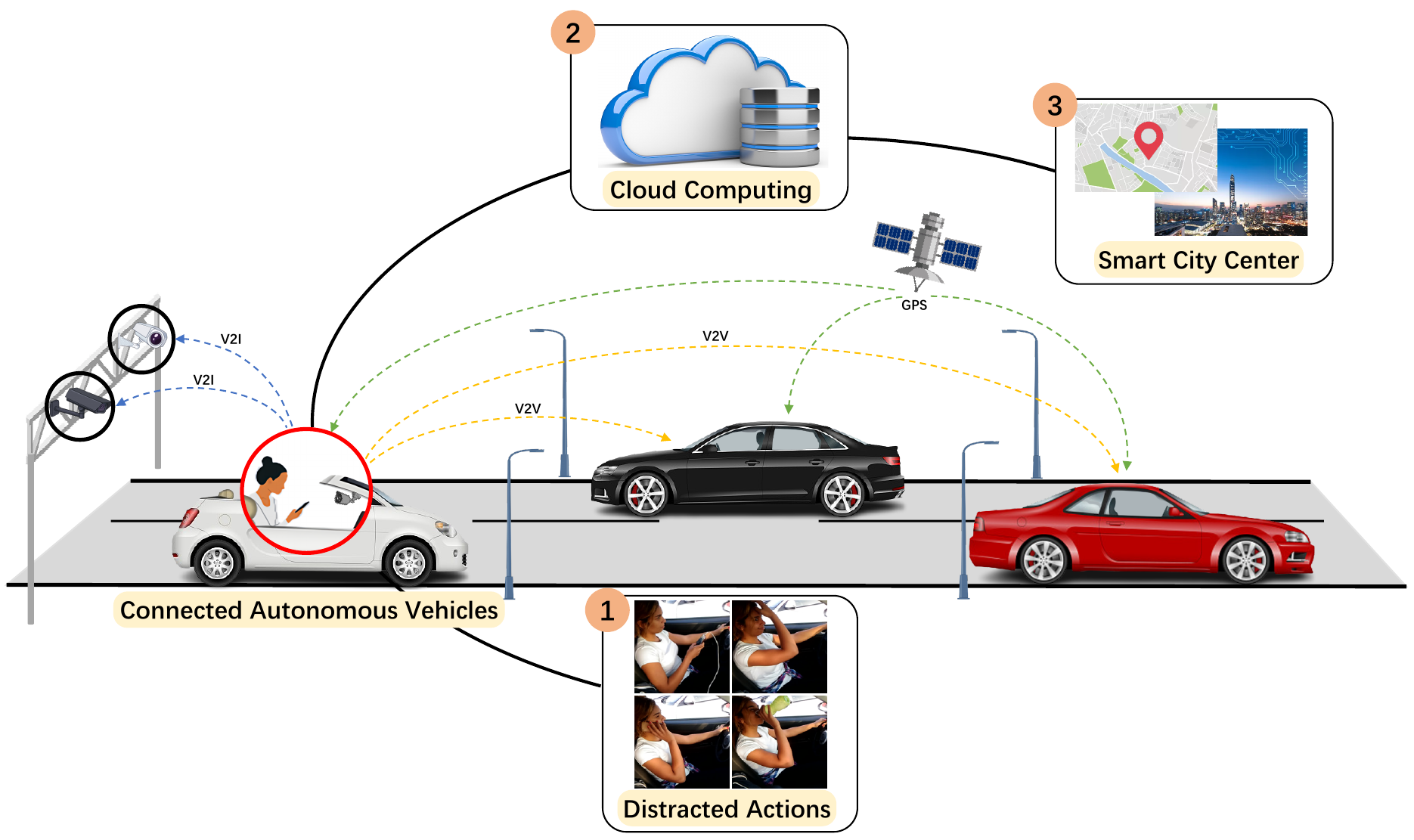}
\caption{Vision-based driver distraction detection employed in intelligent transportation systems. The design of the figure is inspired by \cite{nan-tiv}.}
\label{nan-tiv-fig}
\end{figure*}

As a critical component of smart cities and ITS infrastructure, video surveillance systems are expected to play a vital role in enhancing urban safety and security. With the advent of cloud computing and 5G networks, vision-based driver distraction detection algorithms can be seamlessly integrated into smart city surveillance frameworks, improving transportation safety and traffic management. Figure \ref{nan-tiv-fig} illustrates a vision-based driver distraction detection system for smart cities, where in-vehicle high-definition cameras monitor driver distraction in real-time. The footage is sent to the cloud for analysis, and the system alerts the smart city center of potential traffic risks. Developing an efficient and accurate driving action recognition algorithm is essential within this framework.

\begin{figure}
\centering
\includegraphics[width=3.4in]{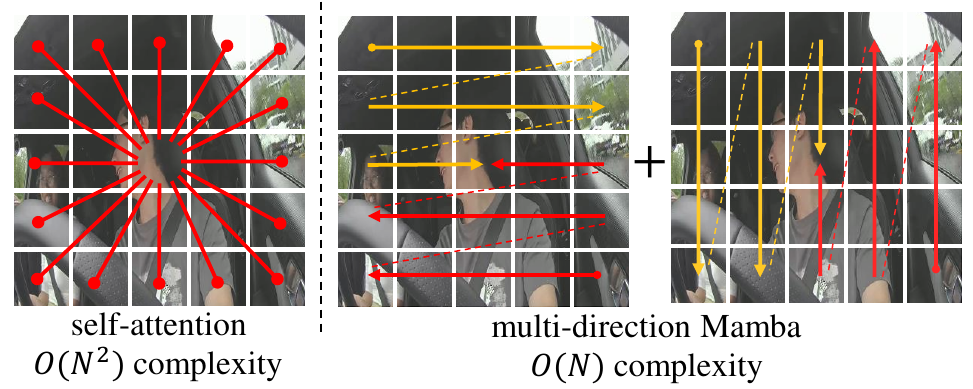}
\caption{Compared to traditional attention mechanisms, the Mamba architecture offers advantages in time complexity.}
\label{show_conplexity}
\end{figure}

Driving action recognition methods include traditional techniques, convolutional neural networks (CNNs), and transformers. Traditional methods, which rely on manually designed features, often struggle with noise in complex scenarios. CNNs, known for their accuracy and real-time performance, are widely used but have limitations in global feature modeling due to their uniform feature extraction. Vision transformers, while surpassing CNNs in image classification and showing promise for driver distraction detection, face challenges with high computational costs and local context extraction. The Mamba structure, a recent innovation in computer vision, excels at extracting global features with linear time complexity, as shown in Figure \ref{show_conplexity}. However, originally designed for long sequences, Mamba has limitations in modeling regional features.

In driver distraction identification, balancing global modeling with local feature extraction is crucial. Overlooking fine-grained details often degrades classification accuracy, highlighting the need for comprehensive feature extraction. Additionally, real-time inference is vital for practical applications, yet many existing algorithms fail to focus on relevant distraction areas, resulting in suboptimal performance, especially on edge devices. To address this, we propose \textbf{D}ual \textbf{S}tate \textbf{D}omain Trans\textbf{Former} (\textbf{DSDFormer}), an innovative Transformer-Mamba network that enhances feature richness by integrating transformers and Mamba, effectively capturing global cues while supporting real-time processing. 

Moreover, public datasets for driver distraction are typically annotated at the video level, often suffering from imprecise or indistinct labels, which hampers high-accuracy classification. This issue has been largely overlooked in the literature. To tackle annotation noise, we introduce \textbf{Temporal Reasoning Confident Learning (TRCL)}, a method that autonomously refines labels by leveraging inter-frame relationships, eliminating the need for manual re-annotation. The principal contributions of this work are as follows:

\begin{itemize}

\item [1)]  
We introduce Temporal Reasoning Confident Learning (TRCL) to address the challenge of imprecise annotations in public datasets. TRCL refines noisy labels by leveraging spatiotemporal continuity and correlations between adjacent frames, offering a more precise and adaptive approach to noise reduction. Extensive evaluations on the AUC-V1 and 100-Driver datasets highlight its effectiveness, significantly improving classification accuracy and label quality over traditional noise-handling techniques.

\item [2)]  
To overcome the high computational demands of transformers and the regional feature extraction limitations of Mamba, we propose the Dual Spatial Domain Attention (DSDA) mechanism. DSDA seamlessly integrates the global modeling strength of transformers with the efficiency of Mamba, enabling precise spatial and state domain feature extraction while maintaining computational efficiency—essential for real-time driver distraction detection.

\item [3)]  
To enhance feature diversity and representation, we design the Spatial-Channel and Multi-Branch Enhancement modules. These modules, leveraging channel attention and depth-wise convolutions, significantly boost the model's capacity to capture both fine-grained spatial details and channel-specific information, addressing the limitations of transformer and Mamba architectures.

\item [4)]  
We present the Transformer-Mamba framework, DSDFormer, which achieves state-of-the-art performance on the AUC-V1, AUC-V2, and 100-Driver datasets. DSDFormer also demonstrates real-time inference on the Nvidia Jetson AGX Orin, making it highly suitable for deployment in intelligent transportation systems, excelling in both accuracy and real-time performance required for practical applications.

\end{itemize}

The rest of the paper is structured as follows: Section \uppercase\expandafter{\romannumeral2} revisits related works, while Section \uppercase\expandafter{\romannumeral3} describes our proposed model architecture and Confident Learning. Section \uppercase\expandafter{\romannumeral4} contains experimental details and results. Section \uppercase\expandafter{\romannumeral5} presents the conclusion of the paper.

\section{Related Works}
\subsection{Traditional Methods}
Earlier studies on driver distraction identification have typically relied on body-specific representations and the extraction of hand-crafted features, such as eye gaze\cite{r1, r2, r3}, facial expression\cite{r4, r5, r6, r7}, head pose\cite{r8, r9, r10, head_pose_tits}, and body pose\cite{r11, r12}. For example, Zhao et al.\cite{r13} employed techniques like homomorphic filtering, skin-like region segmentation, and contourlet transforms to extract driver posture features, followed by Random Forest classification to categorize four driving postures. Seshadri et al.\cite{r14} used the Supervised Descent Method (SDM) to track facial landmarks, extract features, and classify them with a pre-trained classifier. Billah et al.\cite{r17} developed an automatic method to detect and track body parts, using the relative distances between tracking trajectories to extract features, which were then classified using SVM to identify specific distracted behaviors. However, these traditional methods rely heavily on manual feature engineering and are prone to noise interference, often resulting in reduced classification accuracy in practical applications.

\subsection{Convolutional Neural Networks}
Convolutional Neural Networks (CNNs) have shown exceptional performance in computer vision tasks and have become widely adopted for driving action recognition. Significant advances have been made with architectures such as modified VGG\cite{r18, r29}, modified ResNet\cite{r19}, 3D-CNNs\cite{3d_cnn_tcsvt}, and other lightweight models\cite{r20, r22, r23, r26, r28, r30, r33}, enabling accurate and efficient detection of distracted driving behavior, while supporting real-time performance. Several studies have introduced innovative frameworks to further enhance recognition accuracy. For instance, Ardhendu Behera et al.\cite{r25} utilized transfer learning with an adapted DenseNet, while Wu et al.\cite{r31} proposed a Pose-aware Multi-feature Fusion Network that combines global, hand, and body pose features to detect driver hands using posture information. Kose et al.\cite{r32} developed a real-time driver monitoring method using a pre-trained BN-Inception network, which extracts features from sparsely selected action frames and classifies them. However, traditional CNN architectures are inherently limited by their relatively small receptive fields, which restricts their ability to capture fine-grained details in regions critical for distracted driver detection.


\begin{figure*}[!t]
\centering
\includegraphics[width=5.6in]{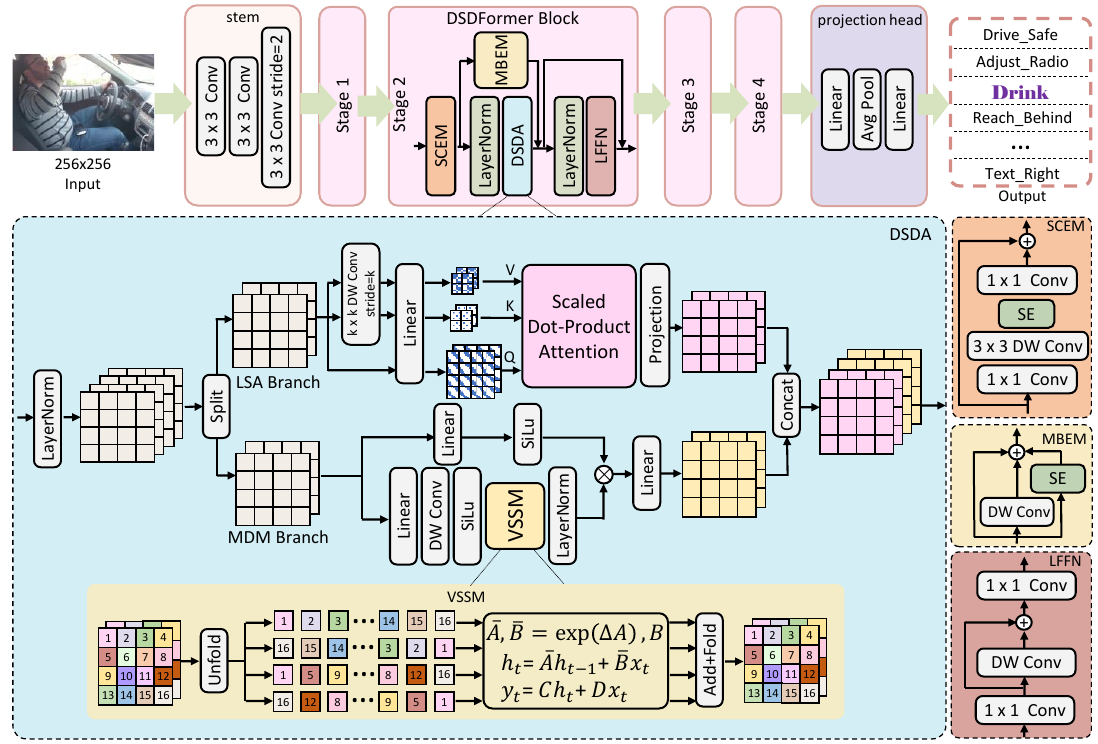}
\caption{DSDFormer comprises the stem, four stages, and the projection head, where the stage stacks several DSDFormer Blocks sequentially. DSDFormer Block consists of a dual state domain attention (DSDA), a spatial-channel enhancement module (SCEM), a multi-branch enhancement module (MBEM), and a lightweight feed-sforward network (LFFN). Conv and DW Conv refers to the convolution and depth-wise convolution, respectively. Linear refers to the fully connected operation, and AvgPool refers to the average pooling operation.}
\label{model_fig}
\end{figure*}

\subsection{Vision Transformer}
Transformers and self-attention mechanisms have revolutionized natural language processing (NLP) and have recently been extended to computer vision tasks. The Vision Transformer (ViT)\cite{ViT} was the first to adapt the pure transformer architecture from NLP to visual tasks, surpassing CNNs in image classification. Transformer-based architectures have since been applied to driver action recognition\cite{driver_transformer1, driver_transformer2, driver_transformer3, driver_transformer4, driver_transformer5, driver_transformer6}. For example, Wharton et al.\cite{r19} incorporated self-attention layers to capture temporal dependencies in video sequences, achieving 84.09\% accuracy on AUC-V1 and 92.50\% on AUC-V2. Yang et al.\cite{r34} proposed the BiRSwinT model, a dual-stream transformer architecture with a feature-level bilinear fusion module, achieving 93.24\% accuracy on AUC-V1. Despite these advancements, transformers in driver distraction recognition still lag behind conventional CNNs. While transformers excel at modeling long-range dependencies, they often underperform in capturing fine-grained local features, which are critical for accurate distracted driving detection. This limitation in feature diversity leads to the loss of essential details and reduced accuracy in fine-grained classification tasks.

\subsection{State Space Models}
State Space Models (SSM) have recently gained attention in natural language processing (NLP) for their ability to efficiently model long sequences with linear time complexity\cite{ssm}. Gu et al.\cite{mamba} introduced the Mamba architecture, a data-driven selective structure SSM that optimizes performance by adapting parameters to input-dependent functions. Building on this, several studies have applied Mamba to computer vision tasks, including classification\cite{vmamba, vision_mamba}, low-level vision tasks\cite{mambair, vmambair, low_level_mamba}, and medical imaging\cite{med_mamba1, med_mamba2, med_mamba3}. Hybrid models that combine Mamba with transformers have also been explored. However, these approaches often either replace attention mechanisms with Mamba\cite{mambair, mamba_replace_atten1}, sacrificing critical regional modeling capabilities, or cascade Mamba with attention mechanisms\cite{mamba_cascade_atten1, mamba_cascade_atten2, mamba_cascade_atten3}, which increases computational overhead. Despite its potential, the application of Mamba in the domain of driver distraction recognition remains largely unexplored.

\subsection{Learning With Noisy Labels}
In computer vision, manual dataset annotation is time-consuming and labor-intensive, often leading to inevitable label noise. To address this issue, early research focused on designing loss functions and regularization techniques to mitigate the impact of noisy labels. Some approaches optimized loss functions by incorporating noise transition matrices\cite{cl_transtrix1, cl_transtrix2}, while others developed robust loss functions\cite{cl_loss} and regularization strategies\cite{cl_regularizer}. Liu et al.\cite{liu_tao} proposed reweighting the loss to ensure better alignment with correct labels. Another research direction explored semi-supervised methods to improve noise detection, with studies employing mentor networks to identify low-loss samples as "clean" data for student networks\cite{cl_teacher1, cl_teacher2}. In driver distraction recognition, the inherent spatiotemporal continuity and action correlation among labeled samples offer an opportunity to address label noise more effectively. However, current methods typically assume that labeled samples are independently and identically distributed, overlooking these correlations and limiting their ability to accurately detect noisy labels in this domain.

\section{Method}
\subsection{Overall Architecture}

We construct a dual state domain transformer, DSDFormer, which integrates both transformers and Mamba for effective long-range modeling and global dependency establishment. As shown in Figure \ref{model_fig}, the stem reduces the input image size with a stride-2 Conv-3×3, followed by two stride-1 Conv-3×3 layers to enhance local information. The model is structured into four stages, each containing multiple DSDFormer blocks for feature transformation. To address the limitation of transformers and Mamba in spatial feature extraction, we introduce channel attention mechanisms in each block to strengthen channel-specific features. The dual state domain attention module is designed to establish global dependencies while reducing computational complexity. Additionally, a multi-branch enhancement structure enriches the diversity of feature representations. A lightweight feed-forward network is used to capture neighboring context more effectively. The model concludes with a projection head that outputs classification results, consisting of a linear layer, global average pooling, and a final linear layer. Detailed analysis of each component within the DSDFormer block is provided in Section \uppercase\expandafter{\romannumeral3}.B.

In driver distraction identification tasks, most public datasets are annotated at the video level, resulting in a significant number of labels with either insufficiently distinct features or entirely erroneous annotations. To address the impact of such noise on model training, we introduce a novel method called Temporal Reasoning Confident Learning, which performs unsupervised noise cleaning without requiring manual reannotation. A detailed explanation of this method is provided in Section \uppercase\expandafter{\romannumeral3}.C.

\subsection{DSDFormer Block}
The proposed DSDFormer Block consists of a dual state domain attention (DSDA), a spatial-channel enhancement module (SCEM), a multi-branch enhancement module (MBEM), and a lightweight feed-forward network (LFFN), as illustrated in Figure \ref{model_fig}.

{\bf{DSDA:}} While transformers are highly effective at extracting global features, their quadratic time complexity results in significant computational overhead, limiting their application in real-world driver distraction identification tasks. Some research mitigates this by partitioning the feature map into patches for self-attention, which speeds up computation. However, this patch-based approach can lead to the loss of fine-grained details, such as hand and eye movements, which are crucial for detecting distraction behaviors. In contrast, the Mamba structure \cite{mamba, mamba2}, with its linear complexity, offers improved computational efficiency and can extract global features at the pixel level, minimizing detail loss. However, Mamba was originally designed for long sequences and lacks regional feature extraction capabilities. To overcome these limitations, we introduce the Dual State Domain Attention (DSDA) mechanism. By integrating transformer and Mamba modules, DSDA enables efficient feature modeling across both spatial and state domains, enhancing the diversity and completeness of feature extraction while improving inference speed. In DSDA, the input $ \mathbf{X} \in \mathbb{R}^{{HW} \times d}$ is split into two parts, $ \mathbf{X_1} \in \mathbb{R}^{{HW} \times \frac{d}{2}}$ and $ \mathbf{X_2} \in \mathbb{R}^{{HW} \times \frac{d}{2}}$, along the channel dimension, with features extracted in parallel through Multi-Direction Mamba (MDM) and Lightweight Self-Attention (LSA), formulated as follows:
\begin{equation}
\begin{aligned}
\operatorname{DSDA}(\mathbf{X})=\operatorname{Concat}[\operatorname{MDM}(\mathbf{X_1}), \operatorname{LSA}(\mathbf{X_2})]\\
\end{aligned}
\end{equation}

\textit{1) state domain attention:} state space models (SSM) are typically regarded as linear time-invariant systems that map a sequence $ x(t) \in \mathbb{R}$ to a sequence $ y(t) \in \mathbb{R}$ by utilizing a hidden state $ h(t) \in \mathbb{R}^N $. The system can be represented as a linear ordinary differential equation(ODE):

\begin{equation}
\begin{aligned}
&  \operatorname{h}^{\prime}(\operatorname{t})= \operatorname{A}\operatorname{h}(\operatorname{t})+ \operatorname{B}\operatorname{x}(\operatorname{t}) \\
& \operatorname{y}(\operatorname{t})=\operatorname{C}\operatorname{h}(\operatorname{t})+\operatorname{D} \operatorname{x}(\operatorname{t})
\end{aligned}
\end{equation}
where N is the state size, $ \operatorname{A} \in \mathbb{R}^{N \times N}$, $ \operatorname{B} \in \mathbb{R}^{N \times 1}$, $ \operatorname{C} \in \mathbb{R}^{1 \times N}$ and $ \operatorname{D} \in \mathbb{R}$.
To integrate Eq.(2) into pratical computer vision algorithms, we can discretize the SSM through the commonly used method zero-order hold (ZOH), which can be defined as follows:
\begin{equation}
\begin{aligned}
\bar{\operatorname{A}} & =\operatorname{e}^{\Delta \operatorname{A}}, \\
\bar{\operatorname{B}} & =\left(\operatorname{e}^{\Delta \operatorname{A}}-\operatorname{I}\right) \operatorname{A}^{-1} \operatorname{B} \approx\Delta \operatorname{B}\\
\end{aligned}
\end{equation}
where $\Delta$ is the timescale parameter to transform the continuous parameters $\operatorname{A}, \operatorname{B}$ to discrete parameters $\bar{\operatorname{A}}, \bar{\operatorname{B}}$ and Eq.(2) can be rewritten as follows:
\begin{equation}
\begin{aligned}
\operatorname{h}_k & =\bar{\operatorname{A}} \operatorname{h}_{k-1}+\bar{\operatorname{B}} \operatorname{x}_k \\
\operatorname{y}_k & =\operatorname{C} \operatorname{h}_k + \operatorname{D} \operatorname{x}_k
\end{aligned}
\end{equation}

Various inputs correspond to the same parameters in Eq.(4). Recently, Mamba introduced a selective scan mechanism(S6) in which $\bar{\operatorname{B}}$, $\operatorname{C}$, and $\Delta$ are derived from input transformations, endowing S6 with dynamic contextual feature modelling capabilities at the pixel-level. We applied S6 and designed the vision state space models (VSSM), as illustrated in Figure \ref{model_fig}. We flattened the feature into 1D vectors in multiple vertical and horizontal directions, and S6 is used to extract global features with linear time complexity. Based on VSSM, our proposed multi-direction Mamba can be formulated as follows:


\begin{equation}
\begin{aligned}
& \operatorname{MDM}(\mathbf{X})=\operatorname{L}(\mathbf{X})*\operatorname{LN}(\operatorname{VSSM}(\operatorname{DW}(\operatorname{L}((\mathbf{X}))))
\end{aligned}
\end{equation}
where $\operatorname{L(\cdot)}$ and $\operatorname{LN(\cdot)}$ are linear layer and layer normalization, respectively.

\textit{2) Spatial domain attention:} Mamba efficiently models global visual features with linear time complexity, providing computational advantages over transformers. However, unlike the inherent sequential dependencies in long text sequences, driver distraction recognition focuses on semantic features where the exact order of local pixel arrangements is less critical. Mamba's method of flattening images into sequences limits its ability to capture intra-regional features. To address this, we designed a lightweight self-attention mechanism that operates in parallel with MDM. To reduce the computational cost of the original self-attention while improving local relevance, we downscale the spatial dimensions of $\operatorname{K}$ and $\operatorname{V}$ using a stride-k depth-wise Conv-k$\times$k. Thus, $ \operatorname{Q} \in \mathbb{R}^{{HW} \times d}$, $ \operatorname{K} \in \mathbb{R}^{\frac{HW}{k^2} \times d}$, and $ \operatorname{V} \in \mathbb{R}^{\frac{HW}{k^2} \times d}$. The formulation for the proposed lightweight self-attention is as follows:

\begin{equation}
\begin{aligned}
&\operatorname{LSA}(\mathbf{X})=\operatorname{Concat}(\operatorname{head}_0, \operatorname{head}_1 \ldots \operatorname{head}_h)\\
&\operatorname{head}_h=\operatorname{Attention}\left(\operatorname{Q}_h, \operatorname{K}_h, \operatorname{V}_h\right) \\
&\operatorname{Attention}\left(\operatorname{Q}_h, \operatorname{K}_h, \operatorname{V}_h\right)=\operatorname{Softmax}\left(\frac{\operatorname{Q}_h \operatorname{K}_h^T}{\sqrt{{\operatorname{d}_k}}}+\operatorname{B}_h\right)\operatorname{V}_h
\end{aligned}
\end{equation}
where $h$ is the index of attention head and $B_h$ is a learnable parameter.


{\bf{SCEM:}} Driver distraction identification relies heavily on visual features concentrated in specific regions of an image, where accurately interpreting localized information is crucial for detecting driver actions. However, transformers and Mamba primarily focus on extracting global features, often neglecting local correlations. Additionally, channel weights are vital in feature modeling\cite{se}, but traditional multi-head self-attention and vision state space models only compute spatial correlations, leading to the loss of important channel-specific information. To address this, we introduce the Spatial-Channel Enhancement Module (SCEM) within the DSDFormer block to improve feature extraction integrity and diversity. As shown in Figure \ref{model_fig}, SCEM incorporates a depth-wise Conv-3$\times$3 to enhance local context information, while a channel attention mechanism reweights and enriches the feature map. SCEM can be defined as:
\begin{equation}
\label{deqn_ex1a}
\operatorname{SCEM}(\mathbf{X})=\operatorname{Conv}(\operatorname{SE}(\operatorname{DW}(\operatorname{Conv}(\mathbf{X}))))+\mathbf{X}
\end{equation}
where $\operatorname{Conv(\cdot)}$ and $\operatorname{DW(\cdot)}$ are Conv-3$\times$3 and depth-wise Conv-3$\times$3, respectively. $\operatorname{SE(\cdot)}$ is the squeeze-excitation module and can be defined as follow:
\begin{equation}
\operatorname{SE}(\mathbf{X})=\operatorname{FC_{2}}(\operatorname{FC_{1}}(\operatorname{GAP}(\mathbf{X}))) * \mathbf{X}
\end{equation}
where $\operatorname{GAP}(\mathbf{X})=\frac{1}{H W} \sum_{i=1, j=1}^{H, W} \mathbf{X}_{i, j}$ is global average pooling in channel dimension and $\operatorname{FC_{1}(\cdot)}$, $\operatorname{FC_{2}(\cdot)}$ are two consecutively fully connected layers.

{\bf{MBEM:}} \;To further enhance the feature representation in both channel-wise and local contexts, we incorporated the MBEM within the DSDFormer block, paralleling with the MDM and LSA. The module combined a channel attention mechanism and a depth-wise Conv-3$\times$3, improving the multiformity and separability of feature extraction by constructing multiple branches. MBEM can be mathematically expressed as:
\begin{equation}
\operatorname{MBEM}(\mathbf{X})=\operatorname{DW}(\mathbf{X})+\operatorname{SE}(\mathbf{X})+\mathbf{X}
\end{equation}

{\bf{LFFN:}} \;To further reduce computational cost and enhance the extraction of local features, we designed the LFFN, which is applied as follows:

\begin{equation}
\begin{aligned}
    \operatorname{LFFN}(\mathbf{X})&=\operatorname{Conv}(\operatorname{F}(\operatorname{Conv}(\mathbf{X})) \\
    \operatorname{F}(\mathbf{X}) &=\operatorname{DW}(\mathbf{X})+\mathbf{X}
\end{aligned}
\end{equation}

With the four components above, the DSDFormer block can be formulated as:
\begin{align}
\mathbf{Y}_i &= \operatorname{SCEM}(\mathbf{X}_{i-1}) \\
\mathbf{Z}_i &= \operatorname{DSDA}(\operatorname{LN}(\mathbf{Y}_{i})) + \operatorname{MBEM}(\mathbf{Y}_{i}) \\
\mathbf{X}_i &= \operatorname{LFFN}(\operatorname{LN}(\mathbf{Z}_{i})) + \mathbf{Z}_{i}
\end{align}

\begin{figure}
\centering
\includegraphics[width=3.4in]{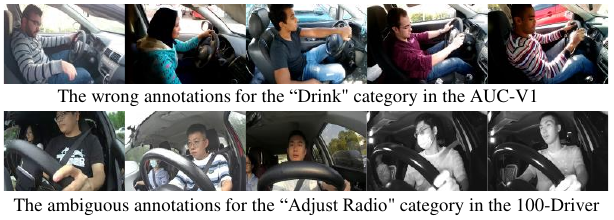}
\caption{Some examples of noisy labels are illustrated.}
\label{noise_sample_fig}
\end{figure}

\subsection{Temporal Reasoning Confident Learning (TRCL)}

In driving action identification, one major challenge is the presence of labels with unclear or inaccurate annotations in video-level datasets (as shown in Figure \ref{noise_sample_fig}), which can significantly degrade the performance of predictive models. Manually re-labelling such data is not only time-consuming and expensive but becomes impractical as dataset sizes grow. This creates a substantial obstacle to achieving high model accuracy, especially when dealing with noisy labels. 

To address this, we introduce Temporal Reasoning Confident Learning (TRCL), an advanced method that builds upon traditional Confident Learning (CL) techniques \cite{cl}. Unlike conventional CL methods, TRCL leverages the temporal continuity inherent in video frames—an aspect often overlooked. By exploiting the natural correlation between consecutive frames, TRCL more effectively identifies and corrects noisy labels, reducing the need for manual re-annotation. This adaptive noise-cleansing process helps overcome the limitations of standard CL methods, improving the overall precision of driving action identification models.

Our method operates on a video-annotated training set \(\mathbb{V} = (v, \tilde{y})^n\), where \(n\) is the number of samples, each potentially associated with noisy labels \(\tilde{y}\). A teacher model predicts probabilities \(\hat{p}\) for each sample across \(m\) classes. For a sample \(v\) labeled \(\tilde{y}=i\), if the predicted probability \(\hat{p}_j(v)\) for another class \(j\) (\(j \neq i\)) exceeds both a threshold \(t_j\) and the probability \(\hat{p}_i(v)\), it suggests that the true label for \(v\) is likely \(y^*=j\). The threshold \(t_j\) is defined as the average predicted probability \(\hat{p}_j(v)\) for all samples labeled \(\tilde{y}=j\):

\begin{equation}
t_j := \frac{1}{\left|\mathbb{V}_{\tilde{y}=j}\right|} \sum_{v \in \mathbb{V}_{\tilde{y}=j}} \hat{p}_j(v)
\label{eqt}
\end{equation}

In this equation, \(\left|\mathbb{V}_{\tilde{y}=j}\right|\) represents the number of samples in \(\mathbb{V}\) with the label \(\tilde{y}=j\).

Next, we build a confusion matrix \(\mathbf{C}_{\tilde{y}, y^*}\) to count the number of samples \(v\) (originally labeled as \(\tilde{y}=i\)) that likely belong to the true label \(y^*=j\):

\begin{equation}
\begin{gathered}
\mathbf{C}_{\tilde{y}=i, y^*=j} := \left|\hat{\mathbb{V}}_{\tilde{y}=i, y^*=j}\right|, \text{ where } \\
\hat{\mathbb{V}}_{\tilde{y}=i, y^*=j} := \{v \in \mathbb{V}_{\tilde{y}=i}: \hat{p}_j(v) \geq t_j, j = \underset{k \in [m]}{\arg \max} \hat{p}_k(v)\}
\end{gathered}
\label{eq11}
\end{equation}

We then normalize \(\mathbf{C}_{\tilde{y}, y^*}\) to create the joint distribution \(\mathbf{Q}_{\tilde{y}, y^*}\):

\begin{equation}
\mathbf{Q}_{\tilde{y}=i, y^*=j} = \frac{\frac{\mathbf{C}_{\tilde{y}=i, y^*=j}}{\sum_{b=1}^m \mathbf{C}_{\tilde{y}=i, y^*=b}} \cdot \left|\mathbb{V}_{\tilde{y}=i}\right|}{\sum_{a, b=1}^m \left(\frac{\mathbf{C}_{\tilde{y}=a, y^*=b}}{\sum_{b=1}^m \mathbf{C}_{\tilde{y}=a, y^*=b}} \cdot \left|\mathbb{V}_{\tilde{y}=a}\right|\right)}
\label{eq12}
\end{equation}



To identify mislabeled samples, we consider four distinct strategies, each leveraging either the confusion matrix \(\mathbf{C}_{\tilde{y}, y^*}\) or the joint distribution \(\mathbf{Q}_{\tilde{y}, y^*}\):

\begin{itemize}
    \item \textbf{Strategy 1:} Samples are flagged as mislabeled if they appear in the off-diagonal elements of \(\mathbf{C}_{\tilde{y}, y^*}\), indicating a discrepancy between predicted and true labels.
    
    \item \textbf{Strategy 2:} For each class \(i\), we select the \(n \cdot \sum_{j \neq i} \mathbf{Q}_{\tilde{y}=i, y^*=j}\) samples with the lowest predicted probability \(\hat{p}_i(v)\), identifying instances where the model exhibits low confidence in the assigned label.
    
    \item \textbf{Strategy 3:} Mislabeled samples are identified by selecting those with the highest difference \(\hat{p}_j(v) - \hat{p}_i(v)\) between predicted probabilities of classes \(i\) and \(j\), using off-diagonal elements of \(\mathbf{Q}_{\tilde{y}, y^*}\) to guide the process.
    
    \item \textbf{Strategy 4:} A hybrid approach combines Strategy 2 and Strategy 3, capturing samples that either display low confidence in the assigned label or exhibit a significant prediction margin between class probabilities.
\end{itemize}

In this study, we opted for strategy 4 to clean the noisy labels, which allows us to derive the set of mislabeled samples, denoted as \(\mathbb{N}\).

\textbf{Temporal Reasoning:} Video data inherently consists of sequential frames, where each frame is temporally correlated with its neighboring frames. This temporal continuity suggests that consecutive frames often share contextual and visual similarities, particularly in scenarios involving continuous actions, such as driving behaviors. To leverage this property, we introduce Temporal Reasoning to enhance the refinement of the mislabeled set \(\mathbb{N}\). Specifically, if a frame \(v_\lambda \in \mathbb{V}\) is identified as mislabeled and reassigned to the true label \(y^*=j\) (\(v_\lambda \in \mathbb{N}_{y^*=j}\)), we exploit the temporal correlation between \(v_\lambda\) and its adjacent frames \(v_{\lambda \pm 1}\) to adjust the predicted probabilities \(\hat{p}\). The adapted probabilities are updated as follows:

\begin{align}
&\hat{p}_j^{\prime}(v_{\lambda \pm 1}) = \hat{p}_j(v_{\lambda \pm 1}) + f(\hat{p}_j(v_{\lambda \pm 1})) \mid v_\lambda \in \mathbb{N}_{y^*=j} \\
&f(\hat{p}_j(v_{\lambda \pm 1})) = \alpha \cdot \hat{p}_j(v_{\lambda \pm 1}) \mid v_\lambda \in \mathbb{N}_{y^*=j}
\end{align}

In the equations above, \(f(\hat{p}_j(v_{\lambda \pm 1}))\) is a scaling function applied to the predicted probability, where \(\alpha\) serves as a weighting factor to modulate the adjustment based on the temporal relationship.

After updating the probabilities for all mislabeled frames in \(\mathbb{N}\), we obtain refined probabilities \(\hat{p}^{\prime}\). Incorporating these refined probabilities into subsequent calculations from Eqs. (14), (15), and (16), and applying the previously discussed identification strategies, we derive a more accurate set of mislabeled samples, denoted as \(\mathbb{N}^{\prime}\).

\section{Experiment}
\subsection{Dataset}
We train and evaluate DSDFormer on the public benchmark, AUC-V1\cite{aucv1}, AUC-V2\cite{aucv2} and 100-Driver\cite{100driver}. AUC-V1 and AUC-V2 are collected from 31 persons and 44 persons, respectively, and both are composed of 10 classes. The 100-Driver dataset comprises 22 categories, 100 persons, and 470,208 images, including daytime and nighttime scenarios. Our empirical analysis revealed that approximately 19\% of the labels in the AUC-V1 dataset were conspicuously erroneous. To more accurately validate our model's effectiveness, we manually curated a gold-standard testing set consisting of 3,570 images. In the 100-Driver dataset, some labels are associated with insufficiently distinct features due to camera angles and steering wheel obstructions. Although these ambiguous labels may impact prediction accuracy, they are not entirely incorrect, and therefore, we opted not to clean the 100-Driver dataset. We validated the effectiveness of the proposed DSDFormer and TRCL methods on the gold-standard AUC-V1 testing set and the original 100-Driver testing set. Additionally, we adhered to the usage protocols of DDT\cite{DDT} and MobileNet+FD\cite{mobilenet_fd} for the AUC-V2 dataset.

\subsection{Implementation Details}
We construct our network based on PyTorch and all experiments are implemented on NVIDIA GeForce RTX4090. We train 200 epochs and employ the AdamW optimizer, the CosineLR strategy, the batch size of 24 and the learning rate of 0.00004.

\begin{table}
    \normalsize
    \belowrulesep=0pt
    \aboverulesep=0pt
    \centering
    \caption{We evaluated the cleaning effect between TRCL and CL on the AUC-V1. TRCL achieves a lower noise rate, remaining noise image number, and higher noise cleaning accuracy.}
    \label{clean_result}
    \begin{tabular}{c c| c c c}
    \toprule
       \small CL & \small TR & \small Remaining Noise & \small Noise(\%) & \small NCA(\%) \\
    \midrule
       \checkmark & $\times$ & 2122 & 17.11 & 69.43 \\
        \checkmark & \checkmark & \textcolor[RGB]{41, 200, 58}{880$\downarrow$} & \textcolor[RGB]{41, 200, 58}{7.99$\downarrow$} & \textcolor{blue}{91.19$\uparrow$} \\
    \bottomrule
        \multicolumn{5}{l}{\scriptsize $Noise=\frac{Remaining \ Noise}{Total \ Noise} \times 100\%$} \\
        \multicolumn{5}{l}{\scriptsize $Noise \ Cleaning \ Accuracy(NCA)=\frac{Noise \ Cleaning \ Num}{Total \ Cleaning \ Num}\times 100\%$} \\
    \end{tabular}
\end{table}

\begin{figure}
\centering
\includegraphics[width=3.4in]{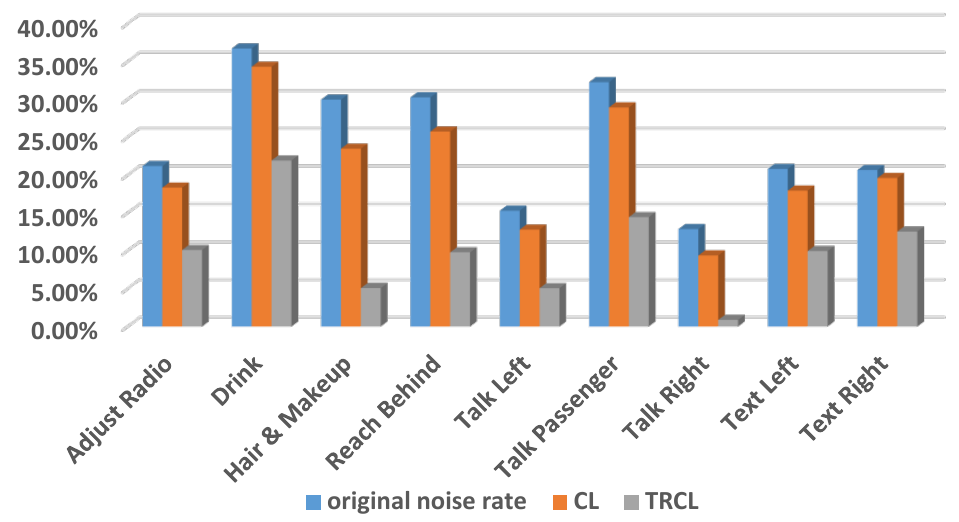}
\caption{We visualized the noise cleaning effect between TRCL and CL on the AUC-V1. TRCL achieves lower noise rates for each category.}
\label{bar_fig}
\end{figure}

\subsection{Evaluation Metrics}
Existing research in driving action identification relies solely on Accuracy for model evaluation. We also utilize Precision, Recall, and F1-score, standard metrics in classification tasks, to provide a more comprehensive comparison. We define the True Positive as TP, False Positive as FP, True Negative as TN, and False Negative as FN, and then the specific evaluation metrics can be formulated as follows.

\textbf{Precision} and \textbf{Recall} can be defined as follows:
\begin{equation}
Pre=\frac{TP}{TP + FP} \times 100\%
\end{equation}

\begin{equation}
Rec=\frac{TP}{TP + FN} \times 100\%
\end{equation}

Considering both Precision and Recall, \textbf{F1-score} evaluates the model performance in a more comprehensive way, which can be defined as follows:
\begin{equation}
F1=2 \times \frac{Pre \times Rec}{Pre + Rec} \times 100\%
\end{equation}

\textbf{Accuracy} is the proportion of correct predictions among the total number of input samples and can be defined as follows:
\begin{equation}
Acc=\frac{TP + TN}{TP + FP+ TN + FN} \times 100\%
\end{equation}

\subsection{Experiment Results With TRCL}

Through empirical observation, we identified approximately 19\% of label noise in the AUC-V1 dataset. To better evaluate the effectiveness of the proposed TRCL algorithm, we manually curated a gold-standard training set of 10,401 samples, which serves as a benchmark for assessing TRCL's adaptive noise labels cleaning performance. We identified the Drive Safe category as devoid of label noise, and consequently, we concentrated our noise-cleaning efforts on the remaining nine categories. The results of the dataset cleaning are presented in Table \ref{clean_result}. After cleaning, TRCL exhibits a notable noise reduction, achieving a Noise Rate of 7.99\%, 9.12\% lower than CL, and the remaining number of noisy labels is 880, 1242 lower than CL. This substantial reduction effectively curtails the incidence of incorrectly labelled images and improves the annotation quality of the dataset, underscoring the effectiveness of our proposed TRCL noise-cleaning methodology. Furthermore, TRCL also exhibits a higher Noise Cleaning Accuracy (NCA) compared to CL. By harnessing the temporal correlation inherent in video-level annotations, we demonstrated the capacity to accurately and substantially diminish the proportion of noisy labels within the dataset. Figure \ref{bar_fig} provides a visual representation of noise rates for each category, comparing the effectiveness of the TRCL method with CL. TRCL consistently yields lower noise rates and demonstrates strong capabilities in noise reduction for every class compared to CL. 

\begin{table*}[!t]
    \setlength
    \tabcolsep{1.5pt}
    \belowrulesep=0pt
    \aboverulesep=0pt
    \renewcommand\arraystretch{0.7}
    \normalsize
    \centering
    \label{cl_result}
    \caption{The performance comparison of various SOTA models trained on the original noisy training set, the training set cleaned by CL, and the training set cleaned by TRCL. Models exhibit only marginal improvements when cleaning using the CL technique. Notable enhancements are observed across multiple performance metrics for various models after applying the TRCL methodology for dataset cleaning.}
    \begin{tabular}{c |c |c c| c c c c c |c c c c c| c c c c c}
      \toprule
      \multirow{2}*{\scriptsize{Model}} & \multirow{2}*{\scriptsize{Venue}} & \multirow{2}*{\scriptsize{CL}} & \multirow{2}*{\scriptsize{TR}} & \multicolumn{5}{c|}{\scriptsize{AUC-V1}} & \multicolumn{5}{c|}{\scriptsize{100-Driver Day-all}} & \multicolumn{5}{c}{\scriptsize{100-Driver Night-all}} \\
      \cmidrule(lr){5-9} \cmidrule(lr){10-14} \cmidrule(lr){15-19}
      & & & & \scriptsize{Acc(\%)} & \scriptsize{Pre(\%)} & \scriptsize{Rec(\%)} & \scriptsize{F1(\%)} & \scriptsize{Err(\%)} & \scriptsize{Acc(\%)} & \scriptsize{Pre(\%)} & \scriptsize{Rec(\%)} & \scriptsize{F1(\%)} & \scriptsize{Err(\%)} & \scriptsize{Acc(\%)} & \scriptsize{Pre(\%)} & \scriptsize{Rec(\%)} & \scriptsize{F1(\%)} & \scriptsize{Err(\%)}\\
       \midrule
       \multirow{3}*{\scriptsize{CAT-CapsNet}\cite{convnext}} & \multirow{3}*{\scriptsize{TITS 2023}} & $\times$ & $\times$ & \scriptsize{97.82} & \scriptsize{98.11} & \scriptsize{97.63} & \scriptsize{97.87} & \scriptsize{2.18} &
       \scriptsize{73.73} & \scriptsize{74.50} & \scriptsize{71.46} & \scriptsize{72.95} & \scriptsize{26.27} &
       \scriptsize{74.49} & \scriptsize{76.67} & \scriptsize{73.23} & \scriptsize{75.05} & \scriptsize{25.51} \\
       ~ & ~ & \checkmark & $\times$ &
       \scriptsize{98.07} & \scriptsize{98.33} & \scriptsize{97.95} & \scriptsize{98.14} & \scriptsize{1.93} &
       \scriptsize{74.16} & \scriptsize{76.39} & \scriptsize{72.81} & \scriptsize{74.55} & \scriptsize{25.84} &
       \scriptsize{74.97} & \scriptsize{76.25} & \scriptsize{74.40} & \scriptsize{75.31} & \scriptsize{25.03} \\
       ~ & ~ & \checkmark & \checkmark &
       \scriptsize{98.32} & \scriptsize{98.49} & \scriptsize{98.24} & \scriptsize{98.37} & \scriptsize{1.68} &
       \scriptsize{76.86} & \scriptsize{76.98} & \scriptsize{76.25} & \scriptsize{76.61} & \scriptsize{23.14} &
       \scriptsize{76.93} & \scriptsize{77.09} & \scriptsize{76.62} & \scriptsize{76.86} & \scriptsize{23.07} \\
       
       \midrule
       \multirow{3}*{\scriptsize{DDT}\cite{convnext}} & \multirow{3}*{\scriptsize{TIV 2024}} & $\times$ & $\times$ & \scriptsize{97.31} & \scriptsize{97.37} & \scriptsize{97.54} & \scriptsize{97.45} & \scriptsize{2.69} &
       \scriptsize{74.48} & \scriptsize{75.33} & \scriptsize{74.45} & \scriptsize{74.88} & \scriptsize{25.52} &
       \scriptsize{65.76} & \scriptsize{68.89} & \scriptsize{64.00} & \scriptsize{65.35} & \scriptsize{34.24} \\
       ~ & ~ & \checkmark & $\times$ &
       \scriptsize{97.37} & \scriptsize{97.26} & \scriptsize{97.81} & \scriptsize{97.53} & \scriptsize{2.63} &
       \scriptsize{75.22} & \scriptsize{75.85} & \scriptsize{74.95} & \scriptsize{75.40} & \scriptsize{24.78} &
       \scriptsize{66.91} & \scriptsize{67.90} & \scriptsize{65.28} & \scriptsize{66.57} & \scriptsize{33.09} \\
       ~ & ~ & \checkmark & \checkmark & 
       \scriptsize{97.59} & \scriptsize{97.98} & \scriptsize{97.71} & \scriptsize{97.85} & \scriptsize{2.41} &
       \scriptsize{76.60} & \scriptsize{77.12} & \scriptsize{76.54} & \scriptsize{76.83} & \scriptsize{23.40} &
       \scriptsize{68.74} & \scriptsize{68.98} & \scriptsize{67.52} & \scriptsize{68.24} & \scriptsize{31.26} \\
       
       \midrule
       \multirow{3}*{\scriptsize{MobileNet+FD}\cite{convnext}} & \multirow{3}*{\scriptsize{TITS 2024}} & $\times$ & $\times$ & \scriptsize{98.07} & \scriptsize{98.30} & \scriptsize{98.29} & \scriptsize{98.30} & \scriptsize{1.93} &
       \scriptsize{78.10} & \scriptsize{79.84} & \scriptsize{77.66} & \scriptsize{78.74} & \scriptsize{21.90} &
       \scriptsize{70.07} & \scriptsize{73.68} & \scriptsize{69.33} & \scriptsize{71.44} & \scriptsize{29.93} \\
       ~ & ~ & \checkmark & $\times$ &
       \scriptsize{98.18} & \scriptsize{98.56} & \scriptsize{98.06} & \scriptsize{98.31} & \scriptsize{1.82} &
       \scriptsize{78.18} & \scriptsize{79.24} & \scriptsize{77.52} & \scriptsize{78.37} & \scriptsize{21.82} &
       \scriptsize{71.16} & \scriptsize{72.57} & \scriptsize{70.53} & \scriptsize{71.53} & \scriptsize{28.84} \\
       ~ & ~ & \checkmark & \checkmark &
       \scriptsize{98.54} & \scriptsize{98.70} & \scriptsize{98.72} & \scriptsize{98.71} & \scriptsize{1.46} &
       \scriptsize{79.72} & \scriptsize{80.48} & \scriptsize{79.38} & \scriptsize{79.92} & \scriptsize{20.28} &
       \scriptsize{73.57} & \scriptsize{75.69} & \scriptsize{72.53} & \scriptsize{74.07} & \scriptsize{26.43} \\
       
       \midrule
       \multirow{3}*{\scriptsize{RMT}\cite{convnext}} & \multirow{3}*{\scriptsize{CVPR 2024}} & $\times$ & $\times$ & \scriptsize{95.83} & \scriptsize{95.59} & \scriptsize{96.30} & \scriptsize{95.95} & \scriptsize{4.17} &
       \scriptsize{79.62} & \scriptsize{80.35} & \scriptsize{79.60} & \scriptsize{79.97} & \scriptsize{20.38} &
       \scriptsize{74.56} & \scriptsize{74.59} & \scriptsize{73.25} & \scriptsize{73.91} & \scriptsize{25.44} \\
       ~ & ~ & \checkmark & $\times$ &
       \scriptsize{96.02} & \scriptsize{95.84} & \scriptsize{96.47} & \scriptsize{96.15} & \scriptsize{3.98} &
       \scriptsize{80.50} & \scriptsize{81.33} & \scriptsize{80.40} & \scriptsize{80.86} & \scriptsize{19.50} &
       \scriptsize{76.77} & \scriptsize{76.79} & \scriptsize{75.98} & \scriptsize{76.38} & \scriptsize{23.23} \\
       ~ & ~ & \checkmark & \checkmark &
       \scriptsize{96.44} & \scriptsize{96.69} & \scriptsize{96.51} & \scriptsize{96.60} & \scriptsize{3.56} &
       \scriptsize{81.07} & \scriptsize{81.60} & \scriptsize{81.03} & \scriptsize{81.32} & \scriptsize{18.93} &
       \scriptsize{79.03} & \scriptsize{78.49} & \scriptsize{78.28} & \scriptsize{78.39} & \scriptsize{20.97} \\
       
       \midrule
       \multirow{3}*{\scriptsize{TransNext}\cite{convnext}} & \multirow{3}*{\scriptsize{CVPR 2024}} & $\times$ & $\times$ & 
       \scriptsize{96.83} & \scriptsize{96.74} & \scriptsize{97.22} & \scriptsize{96.98} & \scriptsize{3.17} &
       \scriptsize{77.10} & \scriptsize{78.15} & \scriptsize{76.80} & \scriptsize{77.47} & \scriptsize{22.90} &
       \scriptsize{71.50} & \scriptsize{73.40} & \scriptsize{70.61} & \scriptsize{71.98} & \scriptsize{28.50} \\
       ~ & ~ & \checkmark & $\times$ &
       \scriptsize{96.92} & \scriptsize{96.79} & \scriptsize{97.27} & \scriptsize{97.03} & \scriptsize{3.08} &
       \scriptsize{79.32} & \scriptsize{79.93} & \scriptsize{79.20} & \scriptsize{79.56} & \scriptsize{20.68} &
       \scriptsize{72.88} & \scriptsize{74.42} & \scriptsize{72.16} & \scriptsize{73.28} & \scriptsize{27.12} \\
       ~ & ~ & \checkmark & \checkmark &
       \scriptsize{97.40} & \scriptsize{97.59} & \scriptsize{97.58} & \scriptsize{97.58} & \scriptsize{2.61} &
       \scriptsize{80.50} & \scriptsize{81.05} & \scriptsize{80.62} & \scriptsize{80.84} & \scriptsize{19.50} &
       \scriptsize{74.03} & \scriptsize{74.94} & \scriptsize{73.24} & \scriptsize{74.08} & \scriptsize{25.97} \\
       
       \midrule
       \multirow{3}*{\scriptsize{Ours}\cite{convnext}} & \multirow{3}*{\scriptsize{-}} & $\times$ & $\times$ & \scriptsize{98.57} & \scriptsize{98.51} & \scriptsize{98.90} & \scriptsize{98.70} & \scriptsize{1.43} &
       \scriptsize{81.21} & \scriptsize{81.29} & \scriptsize{80.96} & \scriptsize{81.12} & \scriptsize{18.79} &
       \scriptsize{76.93} & \scriptsize{78.21} & \scriptsize{76.45} & \scriptsize{77.32} & \scriptsize{23.07} \\
       ~ & ~ & \checkmark & $\times$ &
       \scriptsize{98.74} & \scriptsize{98.77} & \scriptsize{98.90} & \scriptsize{98.84} & \scriptsize{1.26} &
       \scriptsize{81.75} & \scriptsize{81.90} & \scriptsize{81.44} & \scriptsize{81.67} & \scriptsize{18.25} &
       \scriptsize{77.25} & \scriptsize{77.93} & \scriptsize{77.09} & \scriptsize{77.51} & \scriptsize{22.75} \\
       ~ & ~ & \checkmark & \checkmark &
       \scriptsize \textcolor{blue}{99.02$\uparrow$} & \scriptsize \textcolor{blue}{99.10$\uparrow$} & \scriptsize \textcolor{blue}{99.05$\uparrow$} & \scriptsize \textcolor{blue}{99.07$\uparrow$} & \scriptsize \textcolor[RGB]{41, 200, 58}{0.98$\downarrow$} &
       \scriptsize \textcolor{blue}{83.04$\uparrow$} & \scriptsize \textcolor{blue}{83.08$\uparrow$} & \scriptsize \textcolor{blue}{82.88$\uparrow$} & \scriptsize \textcolor{blue}{82.98$\uparrow$} & \scriptsize \textcolor[RGB]{41, 200, 58}{16.96$\downarrow$} &
       \scriptsize \textcolor{blue}{79.79$\uparrow$} & \scriptsize \textcolor{blue}{80.09$\uparrow$} & \scriptsize \textcolor{blue}{79.86$\uparrow$} & \scriptsize \textcolor{blue}{79.97$\uparrow$} & \scriptsize \textcolor[RGB]{41, 200, 58}{20.21$\downarrow$} \\
      \bottomrule
      \multicolumn{8}{l}{\scriptsize $Err(Error \ rate)=100\% - Acc$} \\
    \end{tabular}
\end{table*}

\begin{table*}
    \setlength
    \tabcolsep{1.3pt}
    \normalsize
    \belowrulesep=0pt
    \aboverulesep=0pt
    \renewcommand\arraystretch{0.65}
    \centering
    \caption{The comparison of relative error reduction between CL and TRCL. We computed relative error reduction based on the prediction accuracy trained on the original training dataset. We can find that the relative error reduction of TRCL significantly improved across various models.}
    \label{relative_error}
    \begin{tabular}{c| c c c c c c| c c c c c c| c c c c c c}
    \toprule
    \multirow{2}*{\fontsize{6pt}{8pt}\selectfont Strategy} & \multicolumn{6}{c|}{\scriptsize AUC-V1 Relative Error Reduction(\%)} & \multicolumn{6}{c|}{\scriptsize 100-Driver Day-all Relative Error Reduction(\%)} & \multicolumn{6}{c}{\scriptsize 100-Driver Night-all Relative Error Reduction(\%)} \\
    \cmidrule(lr){2-7} \cmidrule(lr){8-13} \cmidrule(lr){14-19}
    & \fontsize{6pt}{8pt}\selectfont CAT-CapsNet & \fontsize{6pt}{8pt}\selectfont DDT & \fontsize{5.3pt}{8pt}\selectfont MobileNet+FD & \fontsize{6pt}{8pt}\selectfont RMT & \fontsize{6pt}{8pt}\selectfont TransNext & \fontsize{6pt}{8pt}\selectfont Ours
    & \fontsize{6pt}{8pt}\selectfont CAT-CapsNet & \fontsize{6pt}{8pt}\selectfont DDT & \fontsize{6pt}{8pt}\fontsize{5.3pt}{8pt}\selectfont MobileNet+FD & \fontsize{6pt}{8pt}\selectfont RMT & \fontsize{6pt}{8pt}\selectfont TransNext & \fontsize{6pt}{8pt}\selectfont Ours
    & \fontsize{6pt}{8pt}\selectfont CAT-CapsNet & \fontsize{6pt}{8pt}\selectfont DDT & \fontsize{5.3pt}{8pt}\selectfont MobileNet+FD & \fontsize{6pt}{8pt}\selectfont RMT & \fontsize{6pt}{8pt}\selectfont TransNext & \fontsize{6pt}{8pt}\selectfont Ours \\
    \midrule
       \fontsize{6pt}{8pt}\selectfont CL &
       \fontsize{6pt}{8pt}\selectfont 11.54 & \fontsize{6pt}{8pt}\selectfont 2.08 & \fontsize{6pt}{8pt}\selectfont 5.80 & \fontsize{6pt}{8pt}\selectfont 4.70 & \fontsize{6pt}{8pt}\selectfont 2.66 & \fontsize{6pt}{8pt}\selectfont 11.76 & 
       
       \fontsize{6pt}{8pt}\selectfont 1.66 & \fontsize{6pt}{8pt}\selectfont 2.91 & \fontsize{6pt}{8pt}\selectfont 0.36 & \fontsize{6pt}{8pt}\selectfont 4.30 & \fontsize{6pt}{8pt}\selectfont 9.70 & \fontsize{6pt}{8pt}\selectfont 2.88 &
       
       \fontsize{6pt}{8pt}\selectfont 1.85 & \fontsize{6pt}{8pt}\selectfont 3.34 & \fontsize{6pt}{8pt}\selectfont 3.64 & \fontsize{6pt}{8pt}\selectfont 8.70 & \fontsize{6pt}{8pt}\selectfont 4.84 & \fontsize{6pt}{8pt}\selectfont 1.39 \\
       
       \fontsize{6pt}{8pt}\selectfont TRCL &
       \fontsize{6pt}{8pt}\selectfont \textcolor{blue}{23.08$\uparrow$} & \fontsize{6pt}{8pt}\selectfont \textcolor{blue}{10.42$\uparrow$} & \fontsize{6pt}{8pt}\selectfont \textcolor{blue}{24.64$\uparrow$} & \fontsize{6pt}{8pt}\selectfont \textcolor{blue}{14.77$\uparrow$} & \fontsize{6pt}{8pt}\selectfont \textcolor{blue}{17.70$\uparrow$} & \fontsize{6pt}{8pt}\selectfont \textcolor{blue}{31.37$\uparrow$} &
       
       \fontsize{6pt}{8pt}\selectfont \textcolor{blue}{11.92$\uparrow$} & \fontsize{6pt}{8pt}\selectfont \textcolor{blue}{8.31$\uparrow$} & \fontsize{6pt}{8pt}\selectfont \textcolor{blue}{7.40$\uparrow$} & \fontsize{6pt}{8pt}\selectfont \textcolor{blue}{7.11$\uparrow$} & \fontsize{6pt}{8pt}\selectfont \textcolor{blue}{14.87$\uparrow$} & \fontsize{6pt}{8pt}\selectfont \textcolor{blue}{9.71$\uparrow$} &
       
       \fontsize{6pt}{8pt}\selectfont \textcolor{blue}{9.56$\uparrow$} & \fontsize{6pt}{8pt}\selectfont \textcolor{blue}{8.69$\uparrow$} & \fontsize{6pt}{8pt}\selectfont \textcolor{blue}{11.71$\uparrow$} & \fontsize{6pt}{8pt}\selectfont \textcolor{blue}{17.60$\uparrow$} & \fontsize{6pt}{8pt}\selectfont \textcolor{blue}{8.88$\uparrow$} & \fontsize{6pt}{8pt}\selectfont \textcolor{blue}{12.39$\uparrow$} \\
    \bottomrule
    \multicolumn{8}{l}{\fontsize{6pt}{8pt}\selectfont $Relative \ Error \ Reduction=\frac{Err_{old} - Err_{new}}{Err_{old}} \times 100\%$} \\
    \end{tabular}
\end{table*}

We empirically observed that in the 100-Driver dataset, certain annotated classification features are ambiguous due to camera angles or steering wheel obstructions, making it challenging to precisely identify specific distraction behaviours. These labels, characterized by less distinct behavioural features, may introduce some noise into the model training process, but they cannot be considered entirely incorrect. Therefore, unlike the AUC-V1 dataset, where we manually curated a gold-standard training set to directly demonstrate the effectiveness of TRCL in noise cleansing, we opted to apply the TRCL method directly to cleanse the potentially ambiguous labels and evaluate the resulting improvement in model performance. We validated our proposed TRCL method in the AUC-V1 and 100-Driver datasets across daytime and nighttime scenarios. We conducted a comparative analysis of various SOTA algorithms including CAT-CapsNet\cite{CAT-CapsNet}, DDT\cite{DDT}, MobileNet+FD\cite{mobilenet_fd}, RMT-S\cite{rmt} and TransNext-Base\cite{transnext}. As detailed in Tables \uppercase\expandafter{\romannumeral2} and \ref{relative_error}, the models exhibit only marginal improvements when cleaning using the CL technique. The dataset retains a substantial number of erroneous annotations, thereby adversely affecting the training process. Furthermore, upon applying the TRCL methodology for dataset cleaning, notable enhancements are observed across various datasets for diverse classification models. The error rate of MobileNet+FD on AUC-V1 decreased from 1.93\% to 1.46\%, on Day-all from 21.90\% to 20.28\%, and on Night-all from 29.93\% to 26.43\%, with relative error reduction rates of 24.64\%, 7.40\%, and 11.71\%, respectively. The error rate of our DSDFormer decreased from 1.43\% to 0.98\% on AUC-V1, from 18.79\% to 16.96\% on Day-all, and from 23.07\% to 20.21\% on Night-all, corresponding to relative error reduction rates of 31.37\%, 9.71\%, and 12.39\%, respectively. TRCL leverages the spatiotemporal continuity and action correlations between consecutive frames in video-based annotated datasets to adaptively clean labels that are either significantly erroneous or exhibit insufficient classification features, thereby enhancing the quality of the dataset. The effectiveness of TRCL as a universal training framework across diverse models for handling high-noise datasets is demonstrated, significantly bolstering the models' resilience to annotation noise and enhancing training performance.

\begin{table*}[!t]
    \setlength\tabcolsep{3pt}
    \belowrulesep=0pt
    \aboverulesep=0pt
    \renewcommand\arraystretch{0.7}    
    \normalsize
    \centering
    \label{auv_result}
    \caption{The performance comparison between our proposed DSDFormer and other models, trained on the AUC-V1 and AUC-V2. DSDFormer achieves superior performance across multiple metrics.}
    \begin{tabular}{c |c| c c c c| c c c c| c c c c}
      \toprule
      \multirow{2}*{\scriptsize Model} & \multirow{2}*{\scriptsize Venue} & \multicolumn{4}{c|}{\scriptsize AUC-V1 origin} & \multicolumn{4}{c|}{\scriptsize AUC-V1 clean} & \multicolumn{4}{c}{\scriptsize AUC-V2}\\
      \cmidrule(lr){3-6} \cmidrule(lr){7-10} \cmidrule(lr){11-14}
      &  & \scriptsize Acc(\%) & \scriptsize Pre(\%) & \scriptsize Rec(\%) & \scriptsize F1(\%) & \scriptsize Acc(\%) & \scriptsize Pre(\%) & \scriptsize Rec(\%) & \scriptsize F1(\%) & \scriptsize Acc(\%) & \scriptsize Pre(\%) & \scriptsize Rec(\%) & \scriptsize F1(\%)\\
      \midrule
      \scriptsize CAT-CapsNet & \scriptsize TITS 2023 
       & \scriptsize 97.82 & \scriptsize 98.11 & \scriptsize 97.63 & \scriptsize 97.87 
       & \scriptsize 98.32 & \scriptsize 98.49 & \scriptsize 98.24 & \scriptsize 98.37 
       & \scriptsize 93.05 & \scriptsize 93.35 & \scriptsize 91.92 & \scriptsize 92.01 \\

      \scriptsize DDT & \scriptsize TIV 2024 
       & \scriptsize 97.31 & \scriptsize 97.37 & \scriptsize 97.54 & \scriptsize 97.45 
       & \scriptsize 97.59 & \scriptsize 97.98 & \scriptsize 97.71 & \scriptsize 97.85 
       & \scriptsize 93.59* & \scriptsize - & \scriptsize - & \scriptsize - \\

      \scriptsize MobileNet+FD & \scriptsize TITS 2024 
       & \scriptsize 98.07 & \scriptsize 98.30 & \scriptsize 98.29 & \scriptsize 98.30 
       & \scriptsize 98.54 & \scriptsize 98.70 & \scriptsize 98.72 & \scriptsize 98.71 
       & \scriptsize 94.84* & \scriptsize - & \scriptsize - & \scriptsize -\\

      \scriptsize RMT & \scriptsize CVPR 2024 
       & \scriptsize 95.83 & \scriptsize 95.59 & \scriptsize 96.30 & \scriptsize 95.95 
       & \scriptsize 96.44 & \scriptsize 96.69 & \scriptsize 96.51 & \scriptsize 96.60 
       & \scriptsize 92.34 & \scriptsize 91.42 & \scriptsize 91.68 & \scriptsize 91.55 \\

      \scriptsize TransNext & \scriptsize CVPR 2024 
       & \scriptsize 96.83 & \scriptsize 96.74 & \scriptsize 97.22 & \scriptsize 96.98 
       & \scriptsize 97.40 & \scriptsize 97.59 & \scriptsize 97.58 & \scriptsize 97.58 
       & \scriptsize 93.41 & \scriptsize 93.93 & \scriptsize 91.06 & \scriptsize 92.47 \\

       \scriptsize Ours & \scriptsize - & \textcolor{blue}{\scriptsize 98.57$\uparrow$} & \textcolor{blue}{\scriptsize 98.51$\uparrow$} & \textcolor{blue}{\scriptsize 98.90$\uparrow$} & \textcolor{blue}{\scriptsize 98.70$\uparrow$} 
       & \textcolor{blue}{\scriptsize 99.02$\uparrow$} & \textcolor{blue}{\scriptsize 99.10$\uparrow$} & \textcolor{blue}{\scriptsize 99.05$\uparrow$} & \textcolor{blue}{\scriptsize 99.07$\uparrow$} 
       & \textcolor{blue}{\scriptsize 95.73$\uparrow$} & \textcolor{blue}{\scriptsize 96.00$\uparrow$} & \textcolor{blue}{\scriptsize 95.03$\uparrow$} & \textcolor{blue}{\scriptsize 95.41$\uparrow$} \\
      \bottomrule
      \multicolumn{8}{l}{\scriptsize $*$ indicates that the data is cited from the corresponding paper} \\
    \end{tabular}
\end{table*}

\begin{table*}[!t]
    \setlength\tabcolsep{2.5pt}
    \belowrulesep=0pt
    \aboverulesep=0pt
    \renewcommand\arraystretch{0.7}
    \normalsize
    \centering
    \label{100_driver_result}
    \caption{The performance comparison between our proposed DSDFormer and other models, trained on the 100-Driver Day-all and Night-all. DSDFormer achieves superior performance across multiple metrics.}
    \begin{tabular}{c |c| c c c c| c c c c| c c c c| c c c c}
      \toprule
      \multirow{2}*{\scriptsize Model} & \multirow{2}*{\scriptsize Venue} & \multicolumn{4}{c|}{\scriptsize 100-Driver Day-all origin} & \multicolumn{4}{c|}{\scriptsize 100-Driver Day-all clean} & \multicolumn{4}{c}{\scriptsize 100-Driver Night-all origin} & \multicolumn{4}{c}{\scriptsize 100-Driver Night-all clean}\\
      \cmidrule(lr){3-6} \cmidrule(lr){7-10} \cmidrule(lr){11-14} \cmidrule(lr){15-18}
      &  & \scriptsize Acc(\%) & \scriptsize Pre(\%) & \scriptsize Rec(\%) & \scriptsize F1(\%) & \scriptsize Acc(\%) & \scriptsize Pre(\%) & \scriptsize Rec(\%) & \scriptsize F1(\%) & \scriptsize Acc(\%) & \scriptsize Pre(\%) & \scriptsize Rec(\%) & \scriptsize F1(\%) & \scriptsize Acc(\%) & \scriptsize Pre(\%) & \scriptsize Rec(\%) & \scriptsize F1(\%)\\
      \midrule
       \scriptsize CAT-CapsNet & \scriptsize TITS 2023 
       & \scriptsize 73.73 & \scriptsize 74.50 & \scriptsize 71.46 & \scriptsize 72.95 
       & \scriptsize 76.86 & \scriptsize 76.98 & \scriptsize 76.25 & \scriptsize 76.61 
       & \scriptsize 74.49 & \scriptsize 76.97 & \scriptsize 73.23 & \scriptsize 75.05
       & \scriptsize 76.93 & \scriptsize 77.09 & \scriptsize 76.62 & \scriptsize 76.86 \\

       \scriptsize DDT & \scriptsize TIV 2024 
       & \scriptsize 74.48 & \scriptsize 75.33 & \scriptsize 74.45 & \scriptsize 74.88 
       & \scriptsize 76.60 & \scriptsize 77.12 & \scriptsize 76.54 & \scriptsize 76.83 
       & \scriptsize 65.76 & \scriptsize 68.89 & \scriptsize 64.00 & \scriptsize 66.35
       & \scriptsize 68.74 & \scriptsize 68.98 & \scriptsize 67.52 & \scriptsize 68.24 \\

       \scriptsize MobileNet+FD & \scriptsize TITS 2024 
       & \scriptsize 78.10 & \scriptsize 79.84 & \scriptsize 77.66 & \scriptsize 78.74 
       & \scriptsize 79.72 & \scriptsize 80.48 & \scriptsize 79.38 & \scriptsize 79.92 
       & \scriptsize 70.07 & \scriptsize 73.68 & \scriptsize 69.33 & \scriptsize 71.44
       & \scriptsize 73.57 & \scriptsize 75.69 & \scriptsize 72.53 & \scriptsize 74.07 \\

       \scriptsize RMT & \scriptsize CVPR 2024 
       & \scriptsize 79.62 & \scriptsize 80.35 & \scriptsize 79.60 & \scriptsize 79.97 
       & \scriptsize 81.07 & \scriptsize 81.60 & \scriptsize 81.03 & \scriptsize 81.32 
       & \scriptsize 74.56 & \scriptsize 74.59 & \scriptsize 73.25 & \scriptsize 73.91
       & \scriptsize 79.03 & \scriptsize 78.49 & \scriptsize 78.28 & \scriptsize 78.39 \\

       \scriptsize TransNext & \scriptsize CVPR 2024 
       & \scriptsize 77.10 & \scriptsize 78.15 & \scriptsize 76.80 & \scriptsize 77.47 
       & \scriptsize 80.50 & \scriptsize 81.05 & \scriptsize 80.62 & \scriptsize 80.84 
       & \scriptsize 71.50 & \scriptsize 73.40 & \scriptsize 70.61 & \scriptsize 71.98
       & \scriptsize 74.03 & \scriptsize 74.94 & \scriptsize 73.24 & \scriptsize 74.08 \\

       \scriptsize Ours & \scriptsize - 
       & \textcolor{blue}{\scriptsize 81.21$\uparrow$} & \textcolor{blue}{\scriptsize 81.29$\uparrow$} & \textcolor{blue}{\scriptsize 80.96$\uparrow$} & \textcolor{blue}{\scriptsize 81.12$\uparrow$} 
       & \textcolor{blue}{\scriptsize 83.04$\uparrow$} & \textcolor{blue}{\scriptsize 83.08$\uparrow$} & \textcolor{blue}{\scriptsize 82.88$\uparrow$} & \textcolor{blue}{\scriptsize 82.98$\uparrow$} 
       & \textcolor{blue}{\scriptsize 76.93$\uparrow$} & \textcolor{blue}{\scriptsize 78.21$\uparrow$} & \textcolor{blue}{\scriptsize 76.45$\uparrow$} & \textcolor{blue}{\scriptsize 77.32$\uparrow$}
       & \textcolor{blue}{\scriptsize 79.79$\uparrow$} & \textcolor{blue}{\scriptsize 80.09$\uparrow$} & \textcolor{blue}{\scriptsize 79.86$\uparrow$} & \textcolor{blue}{\scriptsize 79.97$\uparrow$}\\
      \bottomrule
    \end{tabular}
\end{table*}

\begin{table}
    \setlength\tabcolsep{3pt}
    \belowrulesep=0pt
    \aboverulesep=0pt
    \renewcommand\arraystretch{1}
    \normalsize
    \centering
    \caption{Comparison of the parameter sizes and inference speeds of different models on NVIDIA Jetson AGX Orin.}
    \label{infer_result}
    \begin{tabular}{c c c c}
    \toprule
       \small Model & \small FLOPS(G)) & \small Params(M) & \small FPS \\
    \midrule
       \small CAT-CapsNet & - & \small 8.50* & \small 18  \\
       \small DDT & \small 4.36* & \small 21.89* & \small 11  \\
       \small MobileNet+FD & \small 0.33* & \small 2.24* & \small 42  \\
       \small RMT & \small 4.50* & \small 27.00* & \small 13  \\
       \small TransNext & \small 18.40* & \small 89.70* & \small 7  \\
       \small Ours & \small 2.33 & \small 14.09 & \small 22 \\
    \bottomrule
    \multicolumn{4}{l}{\scriptsize $*$indicates that the data is cited from the corresponding paper} \\
    \end{tabular}
\end{table}

\subsection{Experiment Results With DSDFormer Model}
To validate the efficacy of our proposed DSDFormer, we conducted a comparative performance assessment against other SOTA methods. We trained the models on the original AUC-V1, Day-all and Night-all, and the datasets cleaned by TRCL, respectively. Additionally, we conducted comparative experiments on the AUC-V2 dataset. In the interest of fairness, all experiments adhered to identical hyperparameters and data augmentation techniques to facilitate an equitable comparison. Quantitative performance comparisons are presented in Tables \uppercase\expandafter{\romannumeral4} and \uppercase\expandafter{\romannumeral5}. When trained on the original AUC-V1 dataset, our proposed DSDFormer model delivers notable results, with the Acc of 98.57\%, Pre of 98.51\%, Rec of 98.90\%, and F1 of 98.70\%. These outcomes surpass those of other driver distraction identification algorithms. Specifically, the Acc, Pre, Rec, and F1 are higher than DDT by 1.26\%, 1.14\%, 1.36\%, and 1.25\%, respectively. DSDFormer also achieves optimal performance on the AUC-V2 dataset. Furthermore, when trained on the large-scale dataset, our DSDFormer model exhibits outstanding performance, achieving the Acc of 81.21\%, Pre of 81.29\%, Rec of 80.96\%, and F1 of 81.12\% on the original 100-Driver daytime scenario. The Acc surpasses CAT-CapsNet, DDT, and MobileNet+FD by 7.48\%, 6.73\%, and 3.11\%, while Pre exceeds them by 6.79\%, 5.96\%, and 1.45\%, Rec by 9.50\%, 6.51\%, and 3.30\%, and F1 by 8.17\%, 6.24\%, and 2.38\%, respectively. Our model also achieves the best performance on the nighttime subset of the 100-Driver dataset.

These results affirm the SOTA performance of the Transformer-Mamba based framework DSDFormer on small-scale and large-scale, high-noise and low-noise datasets. Furthermore, a significant performance enhancement is observed for all models when trained on the dataset cleaned by TRCL, underscoring TRCL's innovative potential as a solution for training highly accurate models under high-noise conditions.

\begin{table}
    \setlength\tabcolsep{2.5pt}
    \belowrulesep=0pt
    \aboverulesep=0pt
    \renewcommand\arraystretch{0.9}
    \normalsize
    \centering
    \caption{Ablation Study of DSDA, SCEM, MBEM, and LFFN. The FPS values are evaluated on the NVIDIA Jetson AGX Orin.}
    \label{ablation}
    \begin{tabular}{c |c c c c| c c c c}
    \toprule
       \scriptsize Model & \scriptsize DSDA & \scriptsize SCEM & \scriptsize MBEM & \scriptsize LFFN & \scriptsize Acc(\%) & \scriptsize FLOPS(G)) & \scriptsize Params(M) & \scriptsize FPS \\
    \midrule
       &  &  &  &  & \scriptsize 97.39 & \scriptsize 3.93 & \scriptsize 28.01 & \scriptsize 16  \\
       &  \small \checkmark &  &  &  & \scriptsize 97.71 & \scriptsize 2.11 & \scriptsize 12.32 & \scriptsize 28  \\
       &  \small \checkmark & \small \checkmark &  &  & \scriptsize 97.90 & \scriptsize 2.37 & \scriptsize 14.02 & \scriptsize 24  \\
       &  \small \checkmark &  & \small \checkmark &  & \scriptsize 97.99 & \scriptsize 2.29 & \scriptsize 13.78 & \scriptsize 25  \\
       &  \small \checkmark & \small \checkmark & \small \checkmark &  & \scriptsize 98.34 & \scriptsize 2.75 & \scriptsize 14.43 & \scriptsize 20  \\
       \scriptsize Ours &  \small \checkmark & \small \checkmark & \small \checkmark & \small \checkmark & \scriptsize 98.57 & \scriptsize 2.33 & \scriptsize 14.09 & \scriptsize 22 \\
    \bottomrule
    \end{tabular}
\end{table}

The driver distraction identification task in ITS necessitates a certain level of real-time performance, typically requiring the swift detection of distracted driving behaviours and timely alerts. To evaluate this aspect, we tested the inference speed of the proposed DSDFormer on the NVIDIA Jetson AGX Orin, measuring frames per second (FPS). As shown in the Table \ref{infer_result}, DSDFormer achieved an inference speed of 22 FPS on edge computing devices, meeting the real-time requirement (exceeding 20 FPS).

\begin{figure*}
\centering
\includegraphics[width=7.1in]{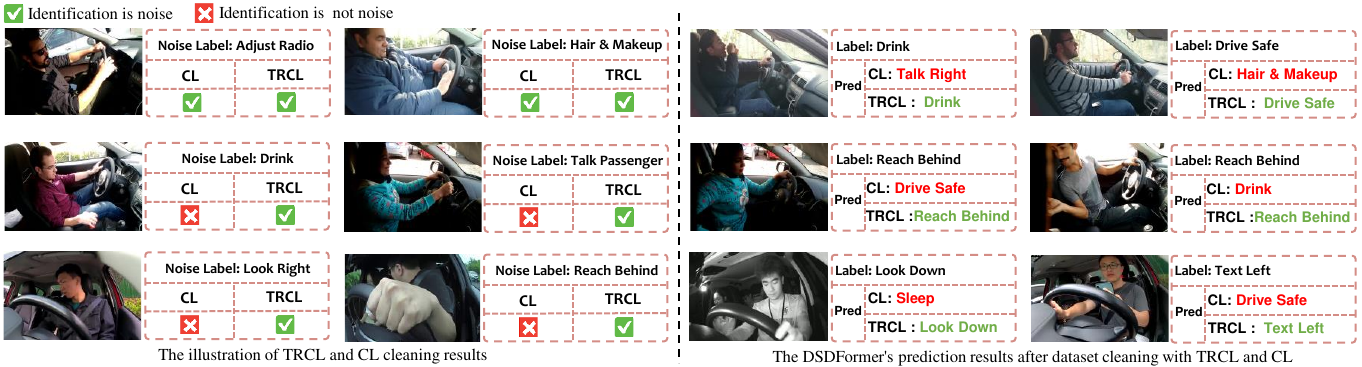}
\caption{Typical illustrative examples of noise cleaning and model prediction. CL specializes in handling prominent noise annotations but fails to identify ambiguous instances. In comparison, TRCL exhibits proficiency in addressing vague noise samples. Furthermore, TRCL mitigates the interference caused by erroneous labels during training, improving prediction accuracy.}
\label{pred_clean_fig}
\end{figure*}

\subsection{Ablation Study}
To assess the effectiveness of the individual modules integrated into our proposed DSDFormer, we conducted ablation experiments on the original AUC-V1 dataset, maintaining consistent hyperparameters to ensure a thorough evaluation. Our model incorporates four distinct modules—DSDA, SCEM, MBEM, and LFFN—each designed to enhance the capture of global and local features while improving inference speed within the DSDFormer block. The outcomes of these ablation experiments are detailed in Table \ref{ablation}. The application of the DSDA module resulted in a 0.32\% increase in accuracy. The inclusion of the SCEM and MBEM modules individually led to additional accuracy improvements of 0.19\% and 0.28\%, respectively. When both the SCEM and MBEM modules were integrated together, accuracy improved by 0.95\%. Furthermore, with the addition of the LFFN module, accuracy reached 98.57\%, reflecting a 1.18\% enhancement over the baseline. Our model, which integrates all four modules within the DSDFormer block, demonstrates competitive performance.


To determine the optimal noise-cleaning approach for driving action recognition, we tested four strategies based on Confident Learning (CL) theory on the AUC-V1 dataset, as summarized in Table \ref{strategy}. Strategy 1 achieved the highest recall (59.54\%), but Strategy 4 exhibited the highest precision (79.24\%), surpassing Strategy 1 by 2.87\%. Given that our primary objective is to minimize false positives in noise identification, we selected Strategy 4 due to its superior precision.

We also explored the influence of the hyperparameter $\alpha$ in the Temporal Reasoning Confident Learning (TRCL) framework, which controls the contribution of temporal context in noise correction. Through systematic evaluation (Table \ref{alpha}), we found that $\alpha = 0.1$ offered the best balance, achieving the highest precision (79.38\%). This setting was adopted as the default in TRCL.

Furthermore, we observed that noise in the AUC-V1 dataset was unevenly distributed across categories. Strategy 4 combined with TRCL effectively reduced noise, particularly in categories with high labeling ambiguity. The consistent performance across all categories validates the robustness of this approach.

\begin{table}
    \normalsize
    \centering
    \caption{Ablation Study of confident learning implementation strategies on the original AUC-V1 dataset. We selected Strategy 4 due to the highest precision of 79.24\%.}
    \label{strategy}
    \begin{tabular}{c c c c c c}
    \toprule
      \small Strategy & \small Acc(\%) & \small Pre(\%) & \small Rec(\%) & \small F1(\%) \\
    \midrule
       \small 1 & \small 84.26 & \small 76.37 & \small 59.54 & \small 66.91\\
       \small 2 & \small 84.77 & \small 78.73 & \small 58.36 & \small 67.03 \\
       \small 3 & \small 84.78 & \small 77.45 & \small 59.08 & \small 67.03 \\
       \small 4 & \small 84.86 & \small 79.24 & \small 57.21 & \small 66.45 \\
    \bottomrule
    \end{tabular}
\end{table}

\subsection{Visualization Analysis}
We presented illustrative examples of noise cleaning in Figure \ref{pred_clean_fig}. Confident Learning (CL) demonstrates its effectiveness in addressing clear cases of erroneous annotations. However, when faced with more ambiguous instances of driver distraction, CL often struggles to accurately identify and resolve noise. For example, scenarios such as the driver's interaction with a mobile phone positioned centrally on the steering wheel present labeling challenges, where it is unclear whether the action should be classified as Text Left or Text Right. In these more nuanced cases, our proposed Temporal Reasoning Confident Learning (TRCL) method, which leverages temporal correlations, effectively identifies and resolves noisy labels. Additionally, we visualized the training outcomes after applying noise cleaning with both TRCL and CL. The dataset cleaned by TRCL exhibited significantly fewer noisy annotations, leading to more accurate predictions of driver distraction behaviors by the model.

\begin{table}
    \normalsize
    \centering
    \caption{Ablation Study of $\alpha$ in TRCL implementation on the original AUC-V1 dataset. We adopted $\alpha=0.1$ as the default setting owing to the highest precision of 79.38\%.}
    \label{alpha}
    \begin{tabular}{c c c c c c c}
    \toprule
      \small $\alpha$ & \small Acc(\%) & \small Pre(\%) & \small Rec(\%) & \small F1(\%) \\
    \midrule
       \small 0.05 & \small 84.96 & \small 78.70 & \small 68.65 & \small 73.33 \\
       \small 0.1 & \small 84.99 & \small 79.38 & \small 67.93 & \small 73.21 \\
       \small 0.15 & \small 84.72 & \small 77.31 & \small 69.75 & \small 73.34 \\
       \small 0.2 & \small 84.69 & \small 76.93 & \small 70.47 & \small 73.56 \\
    \bottomrule
    \end{tabular}
\end{table}

\begin{figure*}
\centering
\includegraphics[width=7.1in]{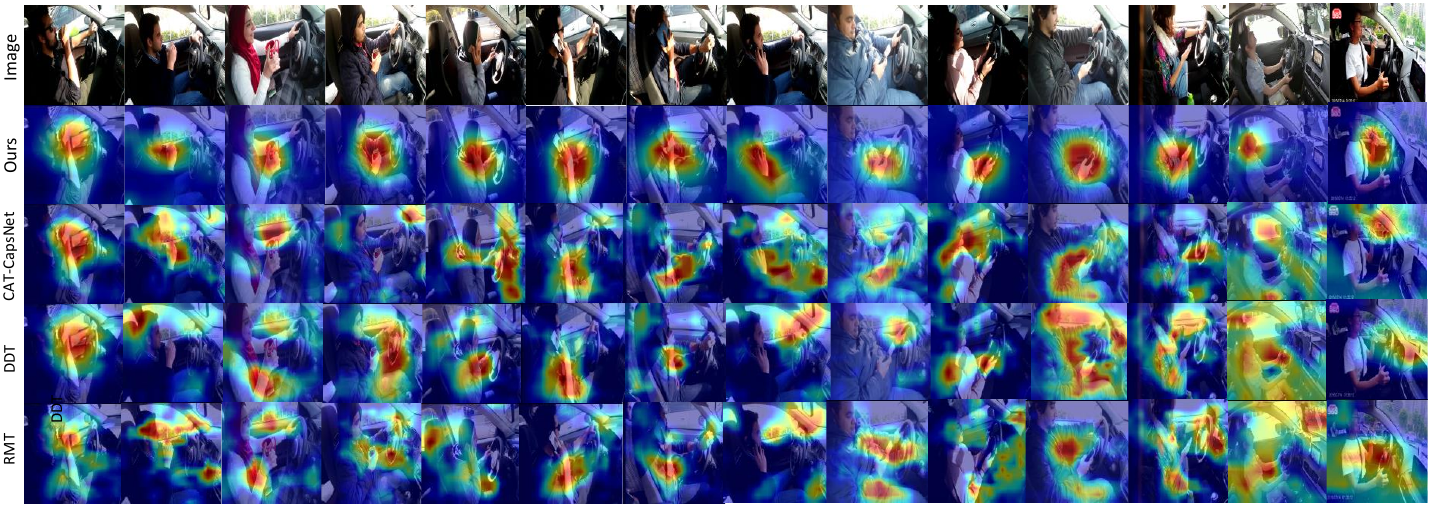}
\caption{Some examples of heat maps for driver distraction identification. From top to bottom are the original images and the prediction results of DSDFormer(Ours), CAT-CapsNet, DDT, and RMT, respectively.}
\label{heatmap_fig}
\end{figure*}

To provide a more comprehensive evaluation of the improvements introduced by DSDFormer, we visualized the performance of various models using heat maps generated by Grad-CAM \cite{grad_cam}, as illustrated in Figure \ref{heatmap_fig}. In these visualizations, brighter areas indicate regions that play a more significant role in driver distraction identification. Existing methods, due to their limited ability to model local features, tend to focus on irrelevant aspects such as the background, driver attire, or facial expressions. In contrast, our proposed model effectively captures both global and local information, as demonstrated by the more precise identification of relevant regions in the heat map visualizations. This enhanced focus on pertinent features contributes to the superior classification accuracy achieved by DSDFormer.

\section{Conclusion}
In this paper, we presented DSDFormer, a novel Transformer-Mamba based framework aimed at enhancing the accuracy and robustness of driver distraction detection. The framework incorporates the Dual State Domain Attention (DSDA) mechanism, which enables the effective capture of both global and local features while ensuring computational efficiency. To further augment feature representation, we introduced Spatial-Channel and Multi-Branch Enhancement modules, addressing the limitations of traditional approaches. Moreover, we proposed Temporal Reasoning Confident Learning (TRCL), an advanced method for refining noisy labels in video-based datasets. Extensive evaluations on the AUC-V1, AUC-V2, and 100-Driver datasets demonstrated that DSDFormer surpasses state-of-the-art models in both accuracy and efficiency, performing well on both edge and cloud platforms.

The findings of this study underscore the potential of DSDFormer as a robust and scalable solution for real-time driver distraction detection, a crucial component in enhancing road safety within intelligent transportation systems. The integration of TRCL not only mitigates the adverse effects of noisy labels but also significantly boosts the model's performance across diverse datasets. Looking ahead, this work can be extended to other computer vision tasks requiring real-time, high-accuracy action recognition. Future research will focus on developing adaptive learning strategies tailored to individual driving patterns, allowing for personalized distraction detection. By incorporating continuous learning and driver-specific data, the system can provide more precise, context-aware predictions, ultimately contributing to greater road safety by adapting to varying driving conditions and behaviors.





\bibliographystyle{IEEEtran}
\bibliography{ref}

\vspace{-9 mm}
\begin{IEEEbiography}[{\includegraphics[width=1in,height=1.25in,clip,keepaspectratio]{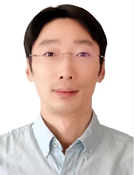}}]
{Junzhou Chen} received his Ph.D. in Computer Science and Engineering from the Chinese University of Hong Kong in 2008, following his M.Eng degree in Software Engineering and B.S. in Computer Science \& Applications from Sichuan University in 2005 and 2002, respectively. Between March 2009 and February 2019, he served as a Lecturer and later as an Associate Professor at the School of Information Science and Technology at Southwest Jiaotong University. He is currently an associate professor at the School of Intelligent Systems Engineering at Sun Yat-sen University. His research interests include computer vision, machine learning, intelligent transportation systems.
\end{IEEEbiography}

\vspace{-9 mm}
\begin{IEEEbiography}[{\includegraphics[width=1in,height=1.25in,clip,keepaspectratio]{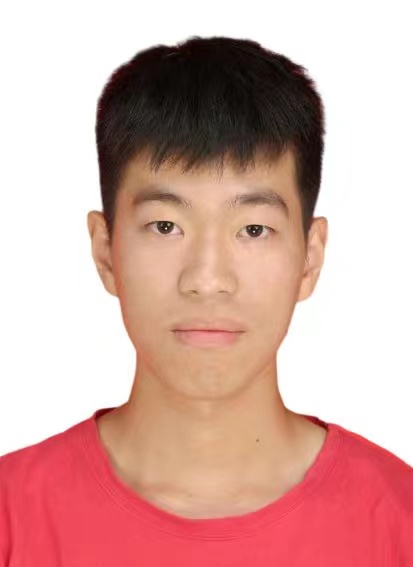}}]
{Zirui Zhang} received the B.S. degree in Automation from Wuhan University of Technology, Wuhan, China, in 2021, and the M.S. degree in Transportation from the School of Intelligent Systems Engineering, Sun Yat-sen University, China, in 2024. Upon completing his graduate studies, he continued his research as a research assistant within the same group, where he has been involved in advancing methodologies in intelligent transportation systems. His research interests include intelligent transportation systems, computer vision, and machine learning, with a particular focus on developing solutions to complex transportation challenges. His work has contributed to several ongoing research projects.
\end{IEEEbiography}

\vspace{-9 mm}
\begin{IEEEbiography}[{\includegraphics[width=1in,height=1.25in,clip,keepaspectratio]{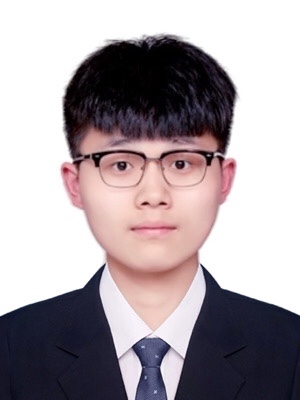}}]
{Jing Yu} received the B.S. degree from Inner Mongolia University, China, in 2022. He is currently pursuing the M.S. degree with the School of Intelligent Systems Engineering, Sun Yat-sen University, China. His research interests include computer vision and intelligent transportation.
\end{IEEEbiography}

\vspace{-9 mm}
\begin{IEEEbiography}[{\includegraphics[width=1in,height=1.25in,clip,keepaspectratio]{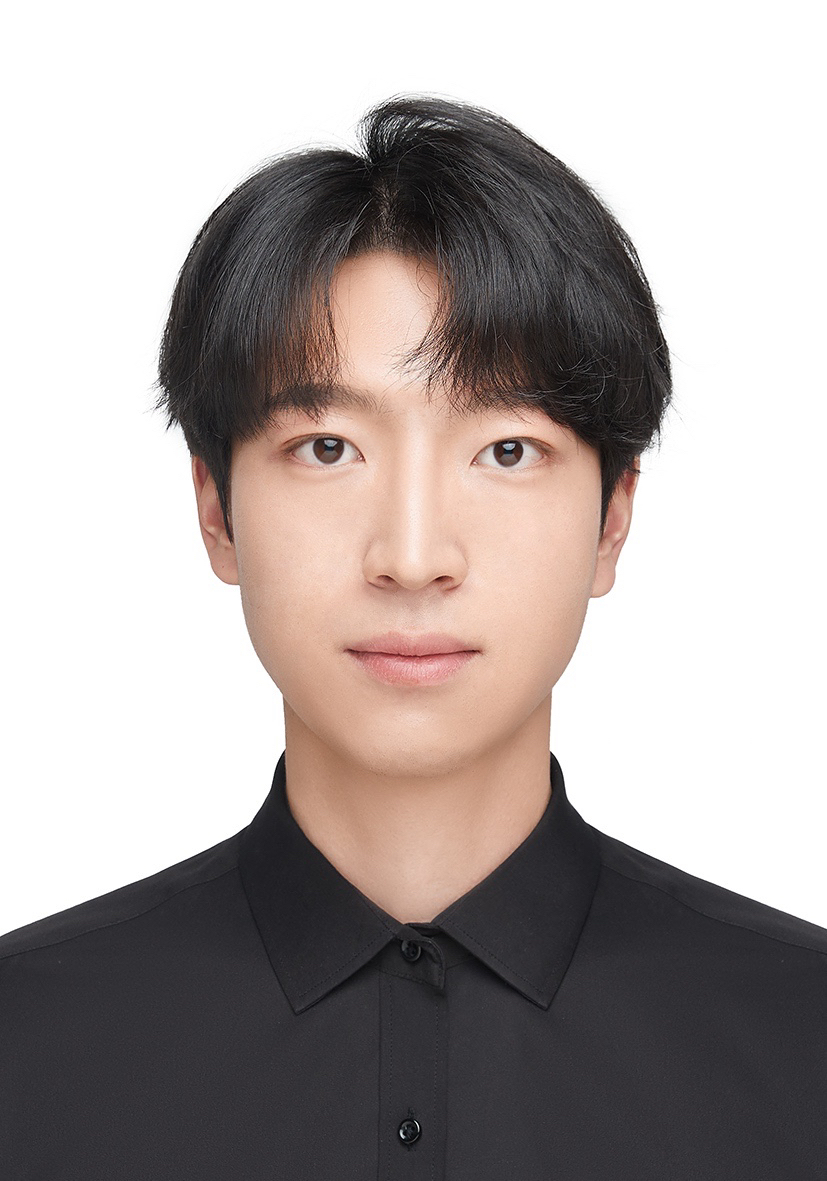}}]
{Heqiang Huang} received the B.S. degree in Electronic Information Science and Technology from Lanzhou University, Lanzhou Province, China in 2023. He is currently working toward the M.S. degree in transportation with the School of Intelligent Systems Engineering, Shenzhen Campus of Sun Yat-sen University, China.
\end{IEEEbiography}

\vspace{-7 mm}
\begin{IEEEbiography}[{\includegraphics[width=1in,height=1.25in,clip,keepaspectratio]{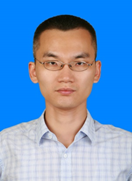}}]
{Ronghui Zhang} received a B.Sc. (Eng.) from the Department of Automation Science and Electrical Engineering, Hebei University, Baoding, China, in 2003, an M.S. degree in Vehicle Application Engineering from Jilin University, Changchun, China, in 2006, and a Ph.D. (Eng.) in Mechanical \& Electrical Engineering from Changchun Institute of Optics, Fine Mechanics and Physics, the Chinese Academy of Sciences, Changchun, China, in 2009. After finishing his post-doctoral research work at INRIA, Paris, France, in February 2011, he is currently an Associate Professor with the Research Center of Intelligent Transportation Systems, School of intelligent systems engineering, Sun Yat-sen University, Guangzhou, Guangdong 510275, P.R.China. His current research interests include computer vision, intelligent control and ITS.
\end{IEEEbiography}

\vspace{-7 mm}
\begin{IEEEbiography}[{\includegraphics[width=1in,height=1.25in,clip,keepaspectratio]{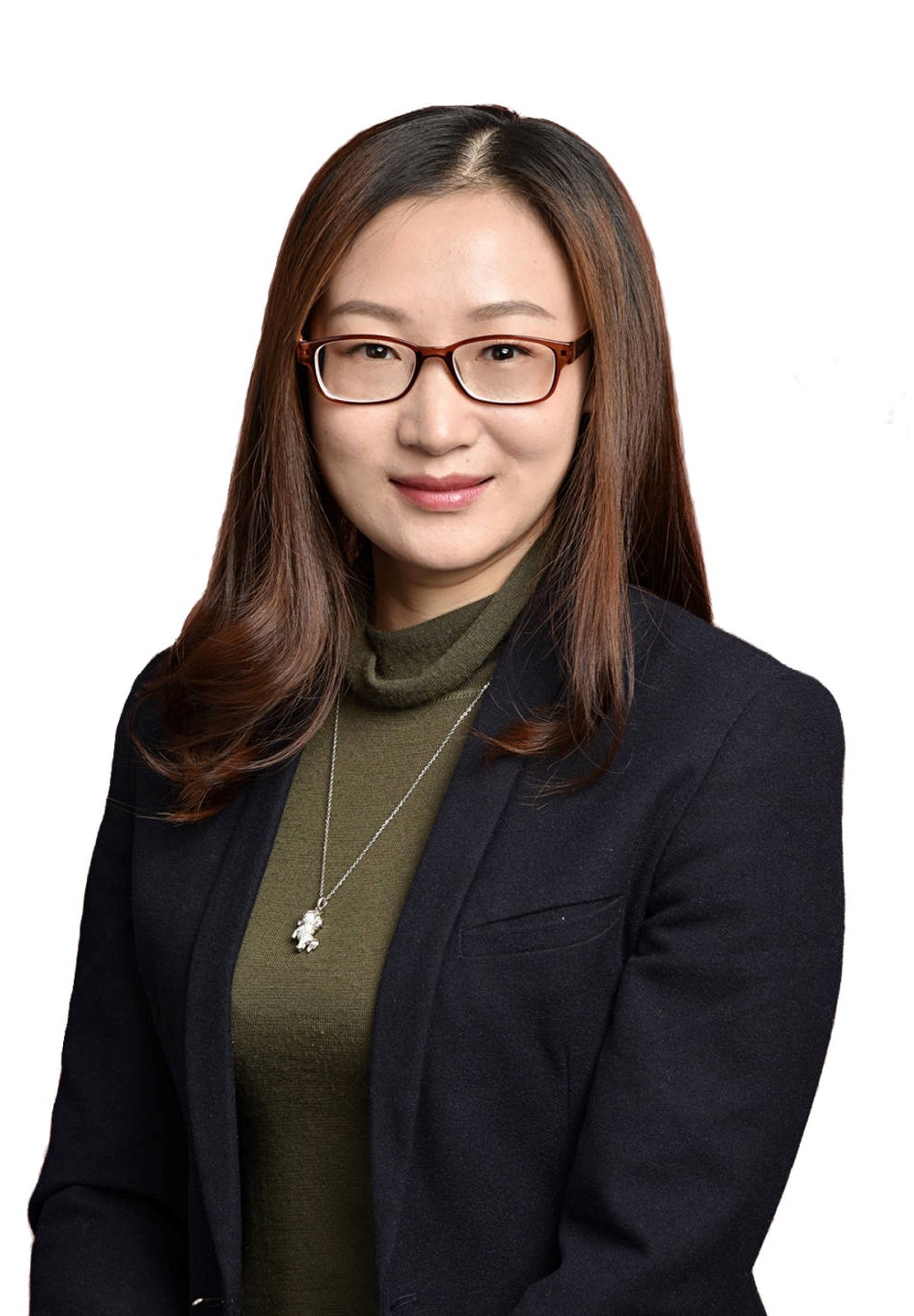}}]
{Xuemiao Xu} received her B.S. and M.S. degrees in Computer Science and Engineering from South China University of Technology in 2002 and 2005 respectively, and Ph.D. degree in Computer Science and Engineering from The Chinese University of Hong Kong in 2009. She is currently a professor in the School of Computer Science and Engineering, South China University of Technology. Her research interests include object detection\&tracking\&recognition, and image\&video understanding and synthesis, particularly their applications in the intelligent transportation system and robots.
\end{IEEEbiography}

\vspace{-7 mm}
\begin{IEEEbiography}[{\includegraphics[width=1in,height=1.25in,clip,keepaspectratio]{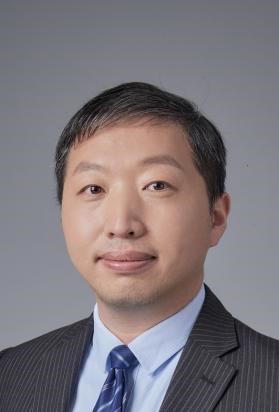}}]
{Bin Sheng} received his BA degree in English and BE degree in computer science from Huazhong University of Science and Technology in 2004, and PhD Degree in computer science from The Chinese University of Hong Kong in 2011. He is currently a full professor in Department of Computer Science and Engineering at Shanghai Jiao Tong University. He serves as the Managing Editor of The Visual Computer, and serves on the Editorial Board of IEEE TCSVT, The Visual Computer, IET Image Processing and J. of Virtual Reality and Intelligent Hardware. In addition, he served as Program Co-Chair of Computer Graphics International (2020-2022), and Conference Co-Chair of Computer Graphics International (2023-2024) and CASA 2024. He also served as the Challenge Co-Chair of DeepDRiD (ISBI2020), DRAC(MICCAI2022) and MMAC(MICCAI2023). He received the Outstanding Contribution Award by Computer Graphics Society in 2023.His research interests include virtual reality, computer graphics and image based techniques. 
\end{IEEEbiography}

\vspace{-9 mm}
\begin{IEEEbiography}[{\includegraphics[width=1in,height=1.25in,clip,keepaspectratio]{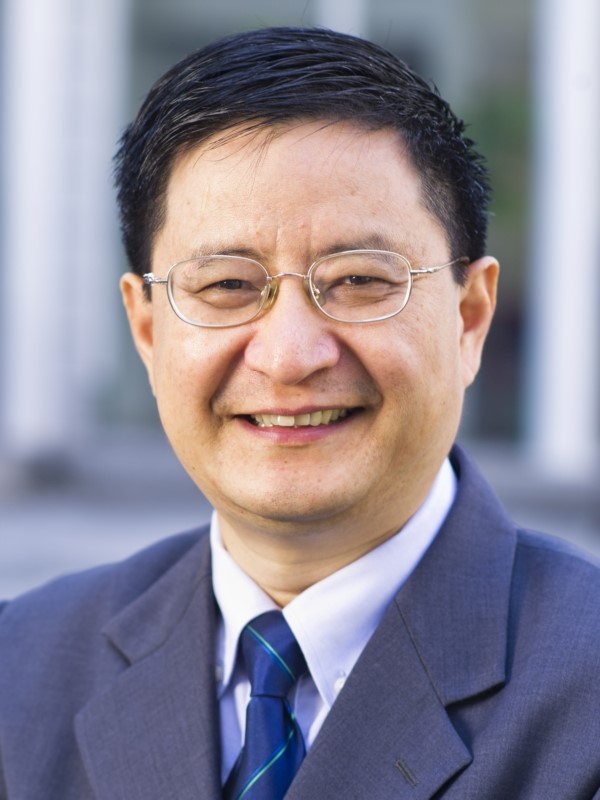}}]
{Hong Yan} received his PhD degree from Yale University. He was Professor of Imaging Science at the University of Sydney and currently is Wong Chun Hong Professor of Data Engineering and Chair Professor of Computer Engineering at City University of Hong Kong. Professor Yan's research interests include image processing, pattern recognition, and bioinformatics. He has over 600 journal and conference publications in these areas. Professor Yan is an IEEE Fellow and IAPR Fellow. He received the 2016 Norbert Wiener Award from the IEEE SMC Society for contributions to image and biomolecular pattern recognition techniques. He is a member of the European Academy of Sciences and Arts and a Fellow of the US National Academy of Inventors.
\end{IEEEbiography}

\vfill

\end{document}